\documentclass[10pt,twocolumn,letterpaper]{article}

\usepackage{cvpr}              

\usepackage{graphicx}
\usepackage{amsmath}
\usepackage{amssymb}
\usepackage{booktabs}

\usepackage[pagebackref,breaklinks,colorlinks]{hyperref}

\usepackage{times}
\usepackage{epsfig}

\usepackage{pifont}
\usepackage{microtype} 
\usepackage{multirow,multicol,booktabs}
\usepackage[flushleft]{threeparttable}
\usepackage{color}
\usepackage{comment}
\usepackage{algorithmic}
\usepackage[ruled,vlined]{algorithm2e}
\usepackage{mwe}
\usepackage{adjustbox}
\usepackage{enumitem}
\usepackage[table]{xcolor}

\newcommand{\cmark}{\text{\ding{51}}}%
\newcommand{\xmark}{\text{\ding{55}}}%

\newcommand{\tablestyle}[2]{\setlength{\tabcolsep}{#1}\renewcommand{\arraystretch}{#2}\centering}

\newcommand{\app}{\raise.17ex\hbox{$\scriptstyle\sim$}}

\newlength\savewidth\newcommand\shline{\noalign{\global\savewidth\arrayrulewidth
  \global\arrayrulewidth 1pt}\hline\noalign{\global\arrayrulewidth\savewidth}}
  
\makeatletter\renewcommand\paragraph{\@startsection{paragraph}{4}{\z@}
  {.5em \@plus1ex \@minus.2ex}{-.5em}{\normalfont\normalsize\bfseries}}\makeatother

\def\fig#1{Fig.~\ref{fig:#1}}

\newcommand{\tb}[3]{\setlength{\tabcolsep}{#2mm}\begin{tabular}{#1}#3\end{tabular}}

\definecolor{ForestGreen}{rgb}{0.13, 0.55, 0.13}
\definecolor{Green}{rgb}{0.0, 0.5, 0.0}
\definecolor{green(munsell)}{rgb}{0.0, 0.66, 0.47}
\definecolor{green(ryb)}{rgb}{0.4, 0.69, 0.2}
\definecolor{green(pigment)}{rgb}{0.0, 0.65, 0.31}

\usepackage{bbm}
\newcolumntype{C}{>{\centering}p{0.05\textwidth}}

\usepackage[ruled,vlined]{algorithm2e}

\definecolor{commentcolor}{RGB}{110,154,155}   
\newcommand{\PyComment}[1]{\ttfamily\textcolor{commentcolor}{\# #1}}  
\newcommand{\PyCode}[1]{\ttfamily\textcolor{black}{#1}} 

\usepackage[capitalize]{cleveref}
\crefname{section}{Sec.}{Secs.}
\Crefname{section}{Section}{Sections}
\Crefname{table}{Table}{Tables}
\crefname{table}{Tab.}{Tabs.}

\def\cvprPaperID{7762} 
\def\confName{CVPR}
\def\confYear{2022}

\begin{document}

\title{Zero-shot and Semi-Supervised Learning with Debiased Pseudo Labeling}
\title{The Devil Is On The Biased Pseudo-Labels: Towards Debiased Pseudo Labeling For Zero-shot and Semi-Supervised Learning}
\title{Towards Debiased Pseudo-labeling \\ For Zero-shot and Semi-Supervised Learning}
\title{Debiased Learning from Naturally Imbalanced Pseudo-Labels}

\author{
\tb{@{}cccc@{}}{7}{
Xudong Wang$^{1}$ & 
Zhirong Wu$^{2}$ & 
Long Lian$^{1}$ & 
Stella X. Yu$^{1}$ 
}\\
\tb{@{}cc@{}}{5}{
$^{1}$UC Berkeley / ICSI &  
$^{2}$Microsoft Research
}}

\maketitle

\begin{abstract}
Pseudo-labels are confident predictions made on unlabeled target data by a classifier trained on labeled source data.  They are widely used for adapting a model to unlabeled data, e.g., in a semi-supervised learning setting.

Our key insight is that pseudo-labels are naturally imbalanced due to intrinsic data similarity, even when a model is trained on balanced source data and evaluated on balanced target data. If we address this previously unknown imbalanced classification problem arising from pseudo-labels instead of ground-truth training labels, we could remove model biases towards false majorities created by pseudo-labels.  

We propose a novel and effective debiased learning method with pseudo-labels, based on counterfactual reasoning and adaptive margins: The former removes the classifier response bias, whereas the latter adjusts the margin of each class according to the imbalance of pseudo-labels.
Validated by extensive experimentation, our simple debiased learning delivers significant accuracy gains over the state-of-the-art on ImageNet-1K: 26\% for semi-supervised learning with 0.2\% annotations and 9\% for zero-shot learning. Our code is
available at: \url{https://github.com/frank-xwang/debiased-pseudo-labeling}.
\end{abstract}
\def\figTeaser#1{
    \captionsetup[sub]{font=small}
    \begin{figure}[#1]
      \centering
      \begin{subfigure}{1.0\linewidth}
        \centering
        \includegraphics[width=0.9\linewidth]{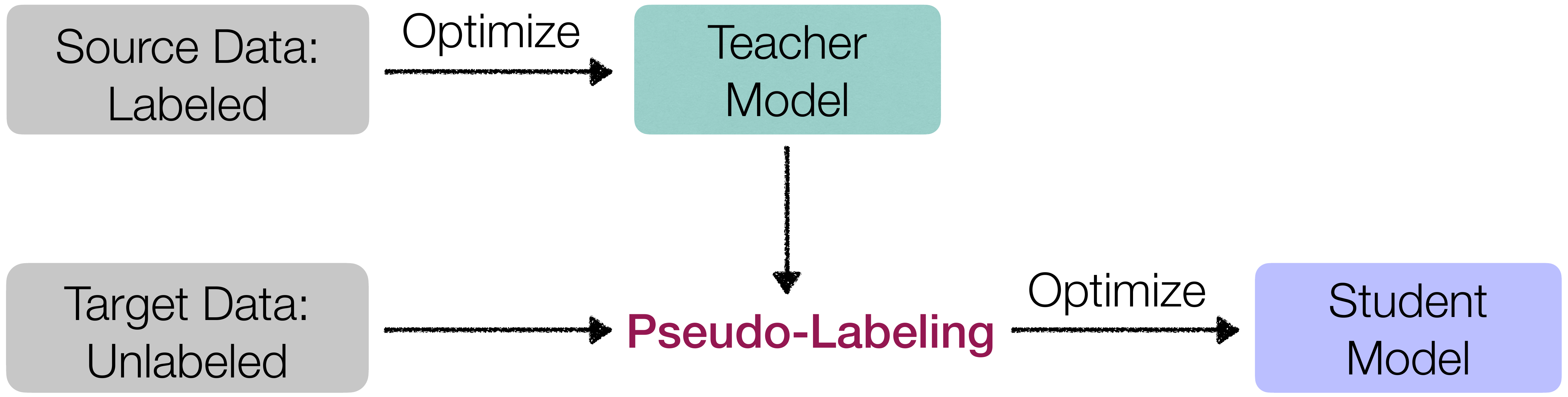}
        \caption{Framework of pseudo-labeling}
        \label{fig:teaser-a}
      \end{subfigure}\vspace{3pt}
      \hfill
      \begin{subfigure}{0.47\linewidth}
        \centering
        \includegraphics[width=0.95\linewidth]{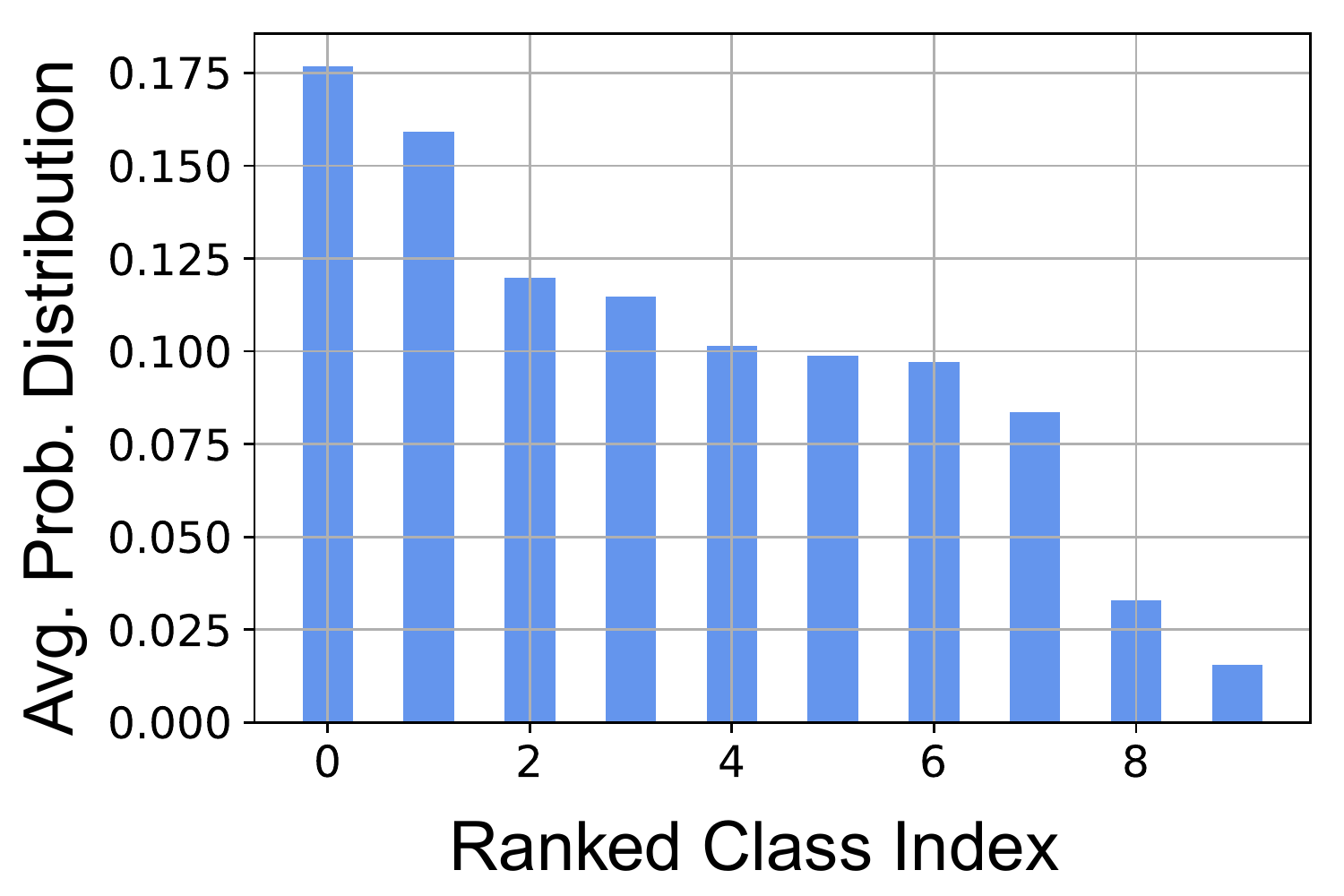}~
        \caption{FixMatch on CIFAR10 SSL}
        \label{fig:teaser-d}
      \end{subfigure}
      \hfill
      \begin{subfigure}{0.47\linewidth}
        \centering
        \includegraphics[width=0.95\linewidth]{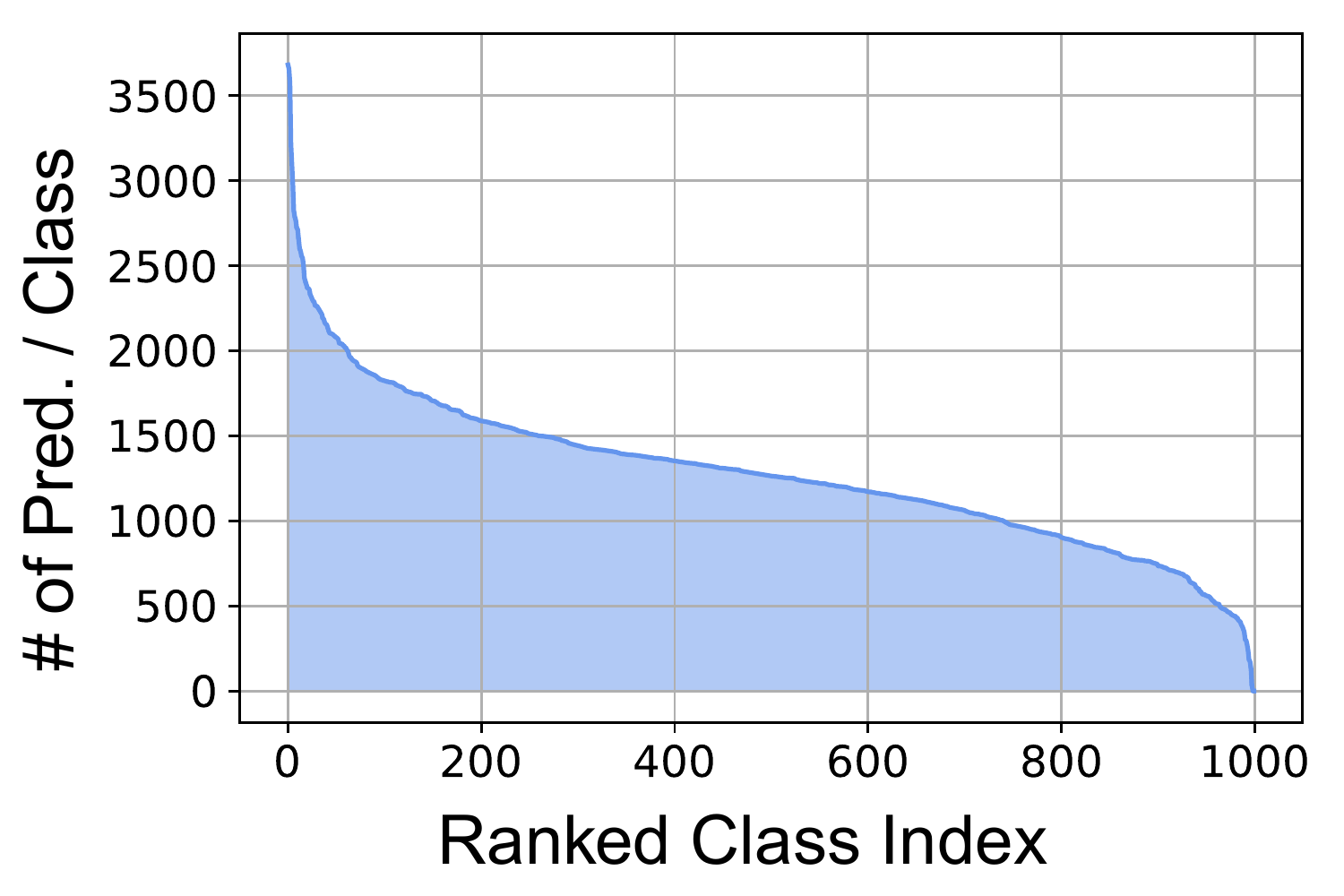}\vspace{-3pt}
        \caption{CLIP on ImageNet ZSL}
        \label{fig:teaser-c}
      \end{subfigure}
      \vspace{-6pt}
      \caption{
      We study the pseudo-labeling-based Semi-Supervised Learning (SSL) and transductive Zero-Shot Learning (ZSL), where both tasks require transferring semantic information learned from labeled source data to unlabeled target data via pseudo-labeling. Surprisingly, we find that pseudo-labels of target data produced by typical SSL and ZSL methods (i.e., FixMatch \cite{sohn2020fixmatch} and CLIP \cite{radford2021learning}) are highly biased, even when both source and target data are class-balanced or even sampled from the same domain.
      }
      \label{fig:teaser-one}
      \vspace{-6pt}
    \end{figure}
}

\figTeaser{t}

\section{Introduction}
\label{introduction}
Real-world observations, as well as non-curated datasets, are naturally long-tail distributed ~\cite{van2018inaturalist, gupta2019lvis}.  Imbalanced classification ~\cite{cao2019learning,kang2019decoupling,wang2021long} tackles such data biases to prevent models from being dominated by head-class instances. Developing visual recognition systems capable of counteracting biases also has significant social impacts ~\cite{mehrabi2021survey}.

While existing methods focus on debiasing from imbalanced ground-truth labels collected by human annotators, we discover that pseudo-labels produced by machine learning models are naturally imbalanced, creating another source for widespread biased learning!

Pseudo-labels are highly confident predictions made by an existing (teacher) model on unlabeled data, which then become part of the training data for supervising the (student) model adaptation to unlabeled data (\fig{teaser-one}a).   When the student model is the teacher model itself, the learning process is also known as {\it self-training} 
\cite{lee2013pseudo, berthelot2019mixmatch, berthelot2019remixmatch, sohn2020fixmatch, xie2020self}.
Pseudo-labeling is widely used in semi-supervised learning (SSL) ~\cite{sohn2020fixmatch,liu2019deep}, domain adaptation ~\cite{na2021fixbi, kang2019contrastive}, and transfer learning~\cite{arnold2007comparative}.

We examine pseudo-label distributions in two common tasks.  {\bf 1)} In zero-shot transfer learning (ZSL) where the source and target domains are different, a pretrained CLIP model~\cite{radford2021learning} produces highly imbalanced predictions on the curated and balanced ImageNet-1K dataset, although the training set of CLIP is approximately balanced (\fig{teaser-one}c). More than 3500 instances are predicted as class 0, 3 times the actual number of samples in class 0.
{\bf 2)} In semi-supervised learning where the source and target domains are the same, FixMatch~\cite{sohn2020fixmatch} trained on labeled CIFAR10 images generates highly biased pseudo-labels on unlabeled images, although both the labeled and unlabeled sets are balanced (\fig{teaser-one}b).

That is, pseudo-labels created by machines are naturally imbalanced, just like ground-truth labels created by humans.  If we address this previously unknown imbalanced classification problem arising from pseudo-labels instead of ground-truth training labels,  we could improve model learning based on pseudo-labels and remove the model bias towards false majorities created by pseudo-labels.

We propose a novel and effective debiased learning method with pseudo-labels, without any knowledge about the distribution of actual classification margins that are readily available to debiased learning with ground-truth labels \cite{kakade2008complexity, liu2016large, wang2018additive}.
It consists of an adaptive debiasing module and an adaptive marginal loss. The former dynamically removes the classifier response bias through counterfactual reasoning,  whereas the latter dynamically adjusts the margin of each class according to the imbalance of pseudo-labels.

Validated by our extensive experiments, our simple debiased learning not only improves the state-of-the-art on ImageNet-1K by 26\% for SSL with 0.2\% annotations and 9\% for ZSL, but is also a universal add-on to various pseudo-labeling methods with more robustness to domain shift.
The imbalanced pseudo-labeling issue is even more severe when the unlabeled raw data is naturally imbalanced, and the model tends to mislabel tail-class samples as head-class. 
By applying debiased learning, we improve SSL performance under long-tailed settings by a large margin.

Our work makes four major contributions.
{\bf 1)} We systematically investigate and discover that pseudo-labels are naturally imbalanced and create biased learning.
{\bf 2)} We propose a simple debiased learning method with pseudo-labeled instances, requiring no knowledge of their actual classification margins.
{\bf 3)} We improve the ZSL/SSL state-of-the-art by a large margin and demonstrate that our debiasing is a universal add-on to various pseudo-labeling models.  
{\bf 4)} We establish a new effective ZSL/SSL pipeline for applying vision-and-language pre-trained models such as CLIP.
\section{Related Work}
\noindent\textbf{Semi-Supervised Learning} 
integrates unlabeled data into training a model given limited labeled data. 
There are four lines of approaches.
{\bf 1)} {Consistency-based regularization} methods impose classification invariance loss on unlabeled data upon perturbations \cite{sajjadi2016regularization, tarvainen2017mean, miyato2018virtual, xie2020unsupervised}. 
{\bf 2)} {Pseudo-labeling} expands model training data from  labeled data to additional unlabeled but confidently pseudo-labeled data \cite{lee2013pseudo, berthelot2019mixmatch, berthelot2019remixmatch, sohn2020fixmatch, xie2020self, li2021comatch}.
{\bf 3)} {Transfer learning}  trains the model first on large unlabeled data through self-supervised representation learning, e.g., contrastive learning, and then on small labeled data through supervised classifier learning \cite{chen2020big,assran2021semi}.
{\bf 4)} {Data-centric SSL}  assumes that labeled data are not given but can be optimally selected among unlabeled data for labeling \cite{wang2021data}.  Focusing on this practical issue of labeled data selection turns out to bring substantial gains for SSL.

CReST \cite{wei2021crest} improves existing SSL methods on class-imbalanced data by leveraging a class-rebalanced sampler, which samples more frequently for the minority class according to the labeled data distribution. CReST does not work when the labeled data is balanced. In contrast, our approach does not assume any prior distribution for the labeled set. 

Although previous literature has achieved tremendous success in SSL, the implicitly biased pseudo-labeling issue in SSL is previously unknown and has not been thoroughly analyzed, which, however, has a great impact on the learning efficiency. The focus of this work is on proposing a simple yet effective debiasing module to eliminate this critical issue.

\noindent\textbf{Zero-shot Classification} refers to the problem setting where a zero-shot model classifies images from novel classes into correct categories that the model has not seen during training \cite{romera2015embarrassingly, pennington2014glove, wang2019survey}. Several strategies have been considered from various sets of viewpoints: 
\textbf{1)} hand-engineered attributes \cite{farhadi2009describing,lampert2013attribute}; \textbf{2)} pretrained embeddings that incorporate prior knowledge in form of semantic descriptions of classes \cite{frome2013devise,socher2013zero}; 
\textbf{3)} modeling relations between seen and unseen classes with knowledge graphs \cite{kampffmeyer2019rethinking,nayak2020zero}; 
\textbf{4)} learning generic visual concepts with vision-language models, allowing zero-shot transfer of the model to a variety of downstream classification tasks \cite{brown2020language,radford2021learning}. 

\noindent\textbf{Long-Tailed Recognition} (LTR)
\label{related-work-long-tailed}
aims to learn accurate ``few-shot” models for classes with a few instances, without sacrificing the performance on ``many-shot" classes, for which many instances are available. 
\textbf{1)} re-balancing/re-weighting method $\tau$-norm \cite{kang2019decoupling} tackles LTR problem by giving more importance to tail classes;
\textbf{2)} margin-based method LDAM \cite{cao2019learning} proposes a label-distribution-aware margin loss to improve the generalization of minority classes by encouraging larger margins for tail classes; 
\textbf{3)} post-hoc adjustment approach modifies a trained model's predictions according to the prior knowledge of class distribution, such as LA \cite{menon2021long}, or pursues the direct causal effect by removing the paradoxical effects of the momentum, such as Causal Norm \cite{tang2020long};
\textbf{4)} ensemble-based approach RIDE \cite{wang2021long} optimizes multiple diversified experts and a dynamic expert routing module to reduce model bias and variance on long-tailed data.

In stark contrast to previous works on LTR which either requires the prior knowledge of class distribution or are applied post-hoc to a trained model, the proposed debias module does not require any prior knowledge and focuses on the biased pseudo-labels issue which is previously unknown.
\def\figInterClassCorrCLIP#1{
    \captionsetup[sub]{font=small}
    \begin{figure*}[#1]
    \centering
    \includegraphics[width=0.95\linewidth]{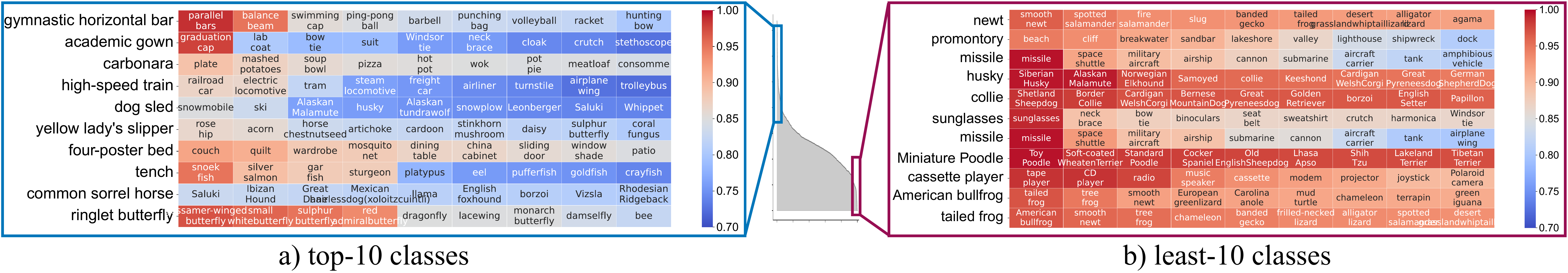}
    \vspace{-6pt}
    \caption{The low-frequency classes of ImageNet, with the least-10 number of CLIP predictions per class, usually have strong inter-class correlations, while the high-frequency classes are the opposite. We compare the cosine similarity between each class's image embedding centroid and embedding centroids of its nine closest ``negative" classes. (better view zoomed in)}
    \label{fig:inter-class-clip}
    \end{figure*}
}

\def\figInterClassCorrFixMatch#1{
    \captionsetup[sub]{font=small}
    \begin{figure}[#1]
      \centering
      \begin{subfigure}{0.47\linewidth}
        \includegraphics[width=1.0\linewidth]{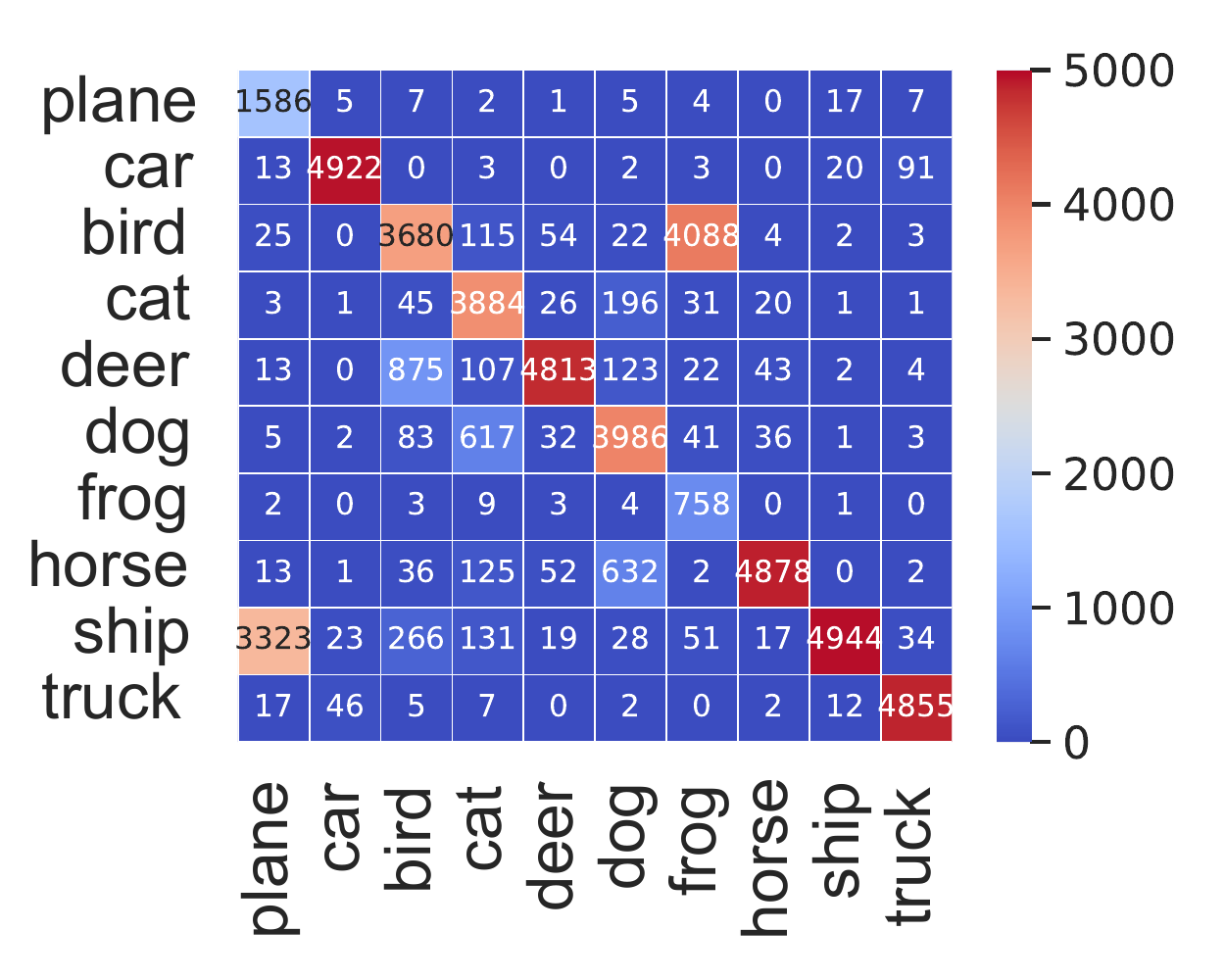}\vspace{-5pt}
        \caption{FixMatch}
        \label{fig:inter-class-fixmatch-a}
      \end{subfigure}
      \hfill
      \begin{subfigure}{0.47\linewidth}
        \includegraphics[width=1.0\linewidth]{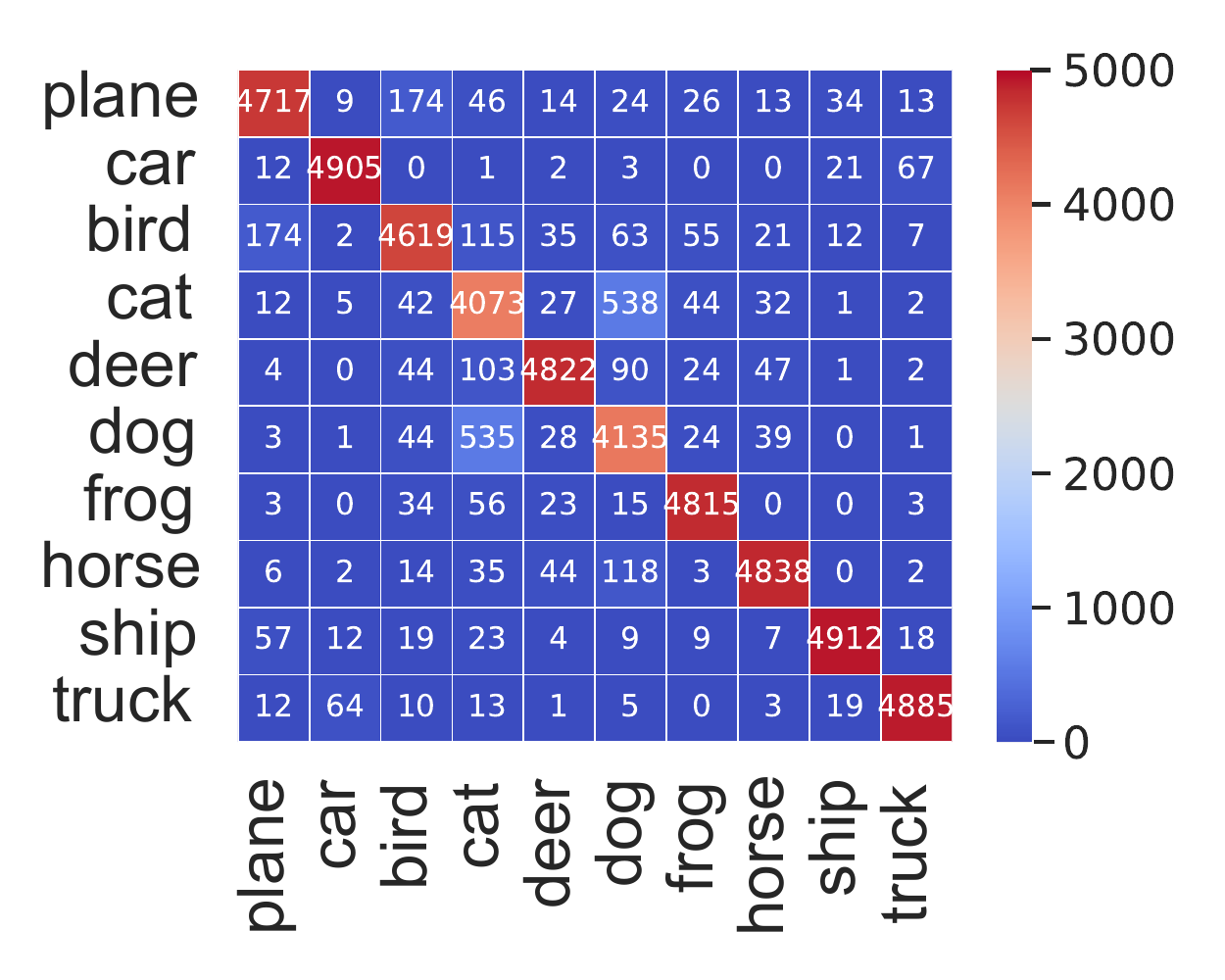}\vspace{-5pt}
        \caption{DebiasPL}
        \label{fig:inter-class-fixmatch-b}
      \end{subfigure}\vspace{-6pt}
      \caption{The cause for pseudo-label biases can be partially attributed to inter-class confounding. For example, FixMatch often misclassifies ``ship" as ``plane". The confusion matrix of FixMatch's and our DebiasPL's pseudo-labels are visualized.}
      \label{fig:inter-class-fixmatch}
    \end{figure}
}

\def\figCLIPBias#1{
    \captionsetup[sub]{font=small}
    \begin{figure}[#1]
      \centering
      \includegraphics[width=0.9\linewidth]{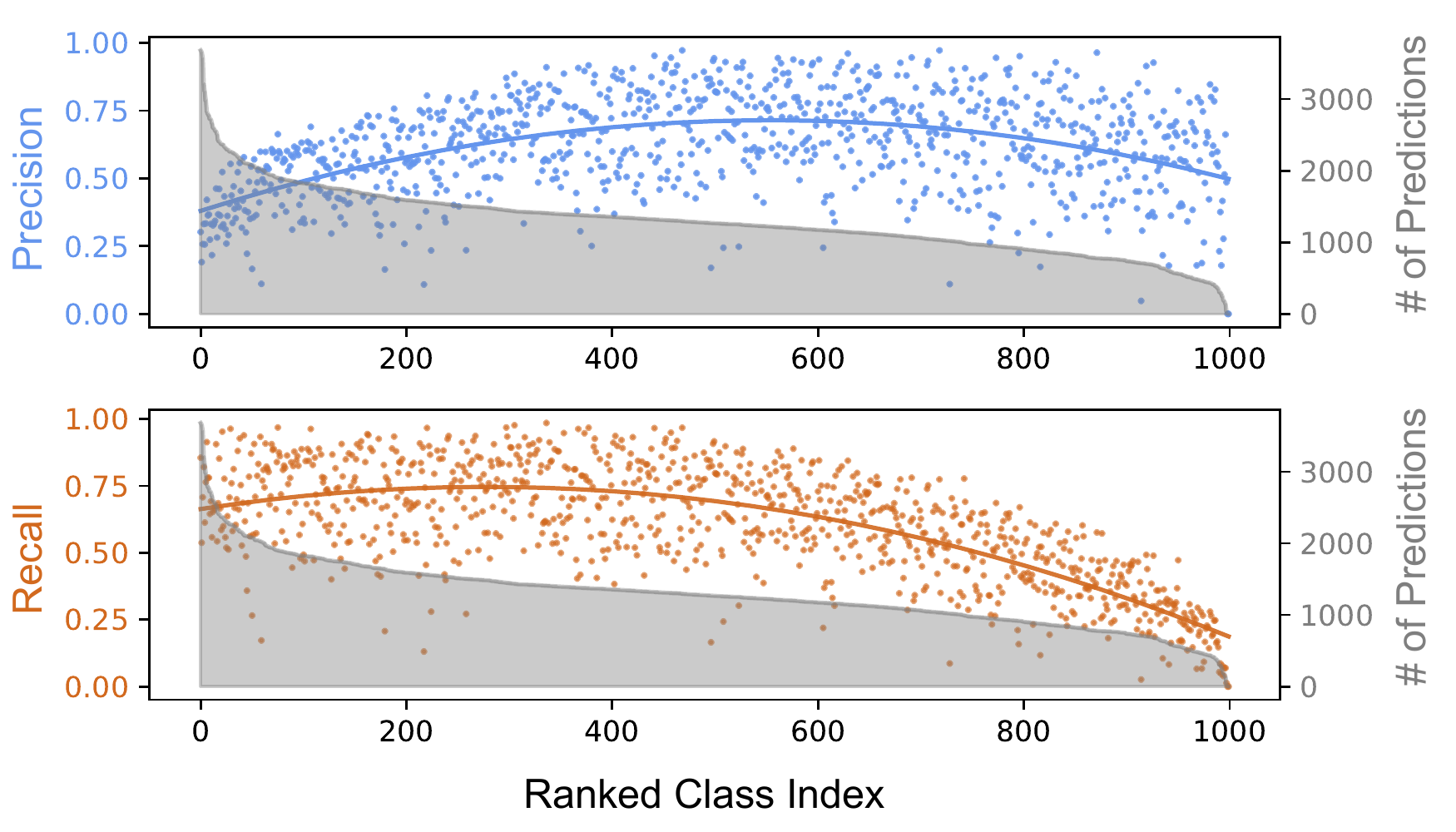}
      \vspace{-6pt}
      \caption{
      Per-class precision and recall of pseudo-label predictions on 1.3M ImageNet instances with a pre-trained CLIP.
      The majority classes with high recall often have less precise pseudo-labels.}
      \label{fig:clip-bias}
    \end{figure}
}

\def\figBiasFixMatch#1{
    \captionsetup[sub]{font=small}
    \begin{figure}[#1]
      \centering
      \includegraphics[width=0.93\linewidth]{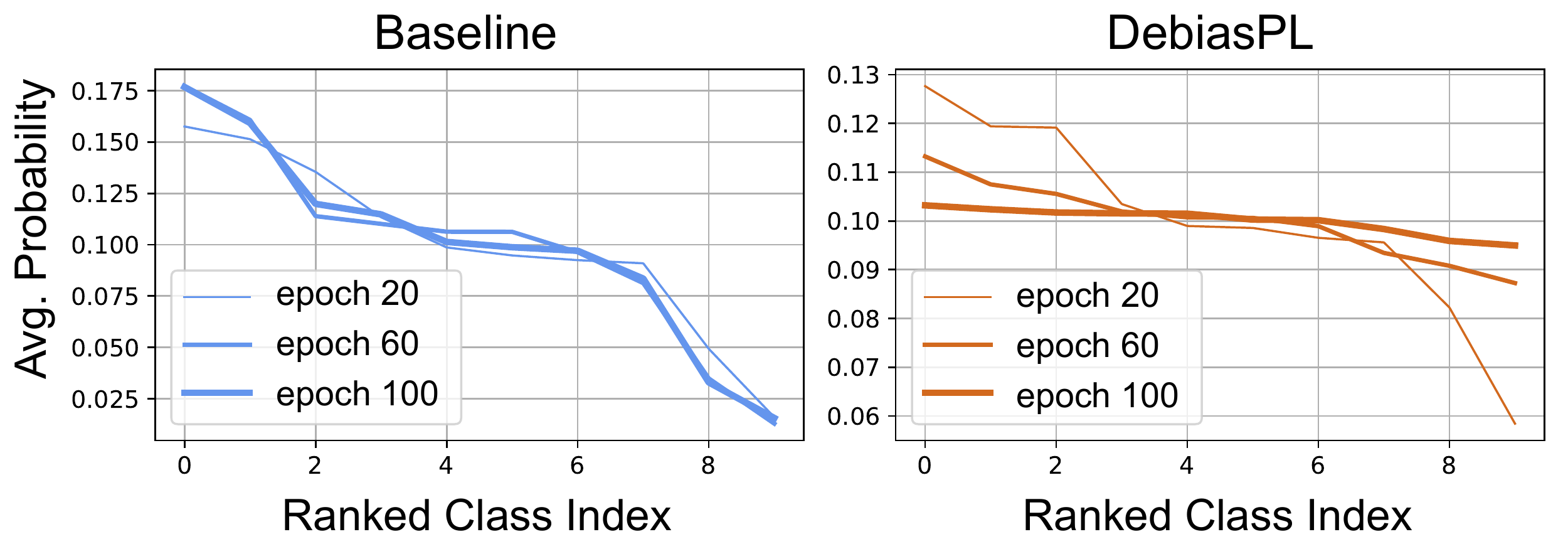}
      \vspace{-6pt}
      \caption{
      FixMatch's pseudo-labels are highly imbalanced across different training stages, even though the unlabeled and labeled data it trains on is class-balanced. In contrast, DebiasPL produces nearly balanced pseudo-labels at late stages. The probability distributions of FixMatch and DebiasPL are averaged over all unlabeled data. The class indices are sorted by average probability. We conduct experiments on CIFAR10 with 4 labeled instances per class.}
    \label{fig:bias-fixmatch}
    \end{figure}
}

\def\figdDiagramDebias#1{
    \captionsetup[sub]{font=small}
    \begin{figure}[#1]
        \centering
        \includegraphics[width=1.0\linewidth]{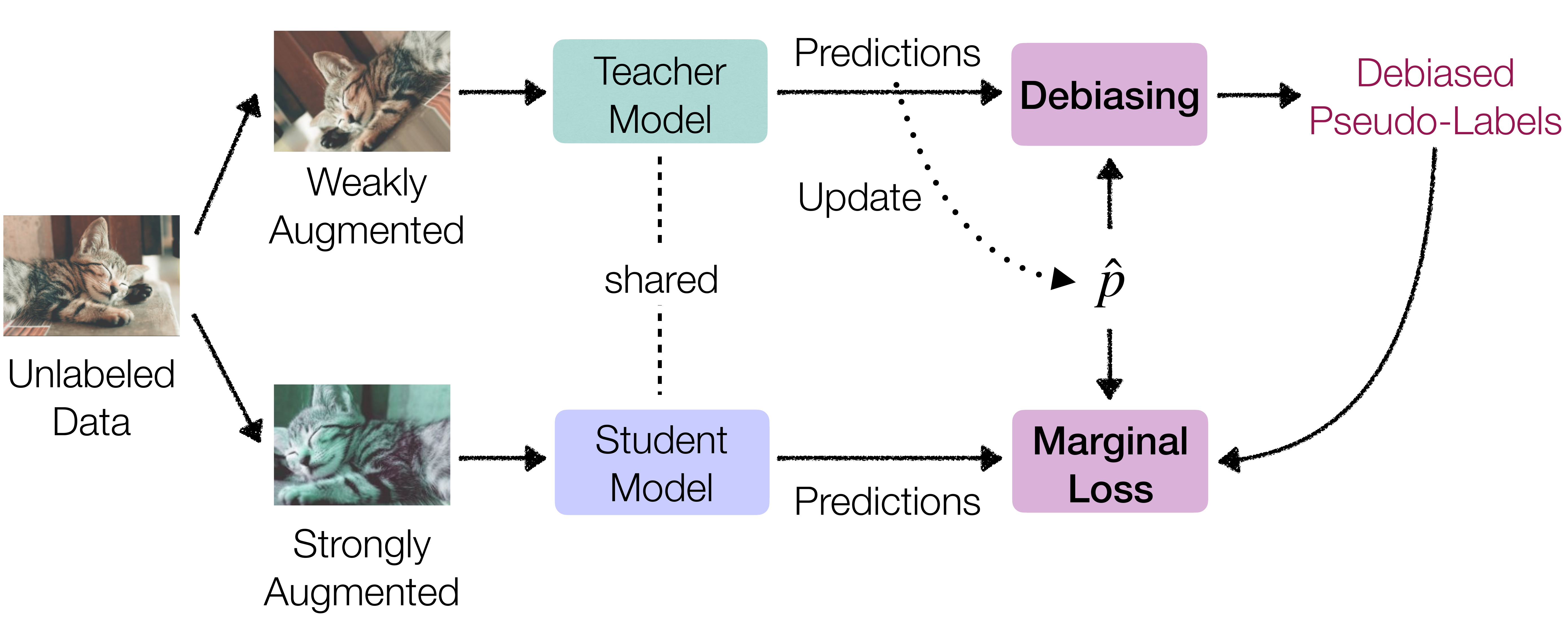}
        \vspace{-16pt}
        \caption{Diagram of the proposed Adaptive Debiasing module and Adaptive Marginal Loss, added to the top of FixMatch.}
        \label{fig:Diagram}
    \end{figure}
}

\def\figdCausal#1{
    \captionsetup[sub]{font=small}
    \begin{figure}[#1]
        \centering
        \includegraphics[width=0.85\linewidth]{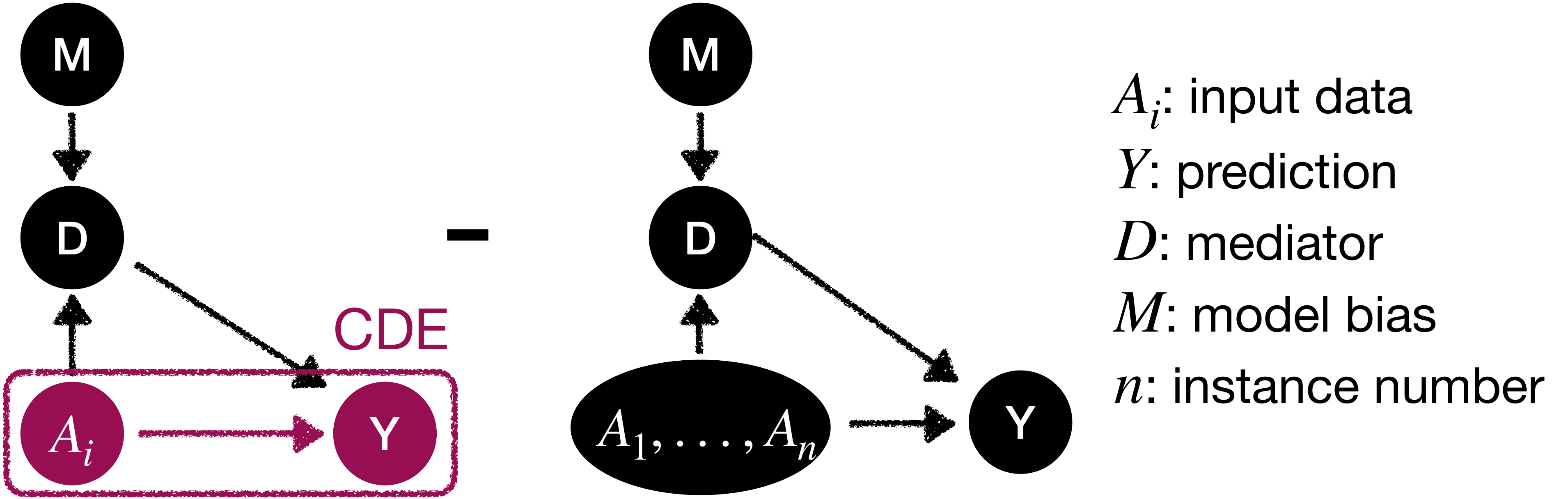}
        \caption{Causal graph of debiasing with counterfactual reasoning.}
        \label{fig:causal-graph}
    \end{figure}
}

\newcolumntype{A}{>{\centering}p{0.03\textwidth}}
\newcolumntype{B}{p{0.04\textwidth}}
\newcommand{\green}[1]{\multicolumn{1}{c}{\color{green(pigment)}#1}}
\newcommand{\red}[1]{\multicolumn{1}{c}{\color{red}#1}}
\newcommand{\cell}[1]{\multicolumn{1}{r}{#1}}

\def\tabDistMethods#1{
    \begin{table}[#1]
    \centering
    \tablestyle{4.5pt}{1.0}
    \small
    \begin{tabular}{l||BBBB}
    \shline
    Desired Properties & \begin{tabular}[c]{@{}c@{}}LA or\\ LDAM \end{tabular} & \begin{tabular}[c]{@{}c@{}}Causal \\ Norm \end{tabular} & \cell{DA} & Ours \\ \hline
    \begin{tabular}[c]{@{}l@{}}Improve representation \\ learning at training time \end{tabular} & \green{$\cmark$} & \red{$\xmark$} & \green{$\cmark$} & \green{$\cmark$} \\ \hline
    \begin{tabular}[c]{@{}l@{}}No prior knowledge on \\ true marginal class distribution \end{tabular} & \red{$\xmark$} & \green{$\cmark$} & \red{$\xmark$} & \green{$\cmark$} \\ \hline
    \begin{tabular}[c]{@{}l@{}}Adaptive as the \\ training progresses \end{tabular} & \red{$\xmark$} & \red{$\xmark$} & \green{$\cmark$} & \green{$\cmark$} \\ \hline
    \begin{tabular}[c]{@{}l@{}}Applicable to both \\ imbalanced and balanced data \end{tabular} & \red{$\xmark$} & \red{$\xmark$} & \green{$\cmark$} & \green{$\cmark$} \\ \hline 
    \begin{tabular}[c]{@{}l@{}}Source and target data can \\ come from varying distributions \end{tabular} & \red{$\xmark$} & \red{$\xmark$} & \red{$\xmark$} & \green{$\cmark$} \\
    \shline
    \end{tabular}\vspace{-4pt}
    \caption{Our method is the only one with all these desired properties. 
    Comparisons with previous works concentrating on resolving training data distribution issues, including LA \cite{menon2021long}, LDAM \cite{cao2019learning}, DA \cite{berthelot2019remixmatch}, Causal Norm \cite{tang2020long} and our DebiasPL, in key properties. \textcolor{green(pigment)}{Desired} (\textcolor{red}{undesired}) properties are in \textcolor{green(pigment)}{green} (\textcolor{red}{red}).
    }
    \label{tab:distMethods}
    \end{table}
}

\def\tabCIFAR#1{
\begin{table*}[#1]
\tablestyle{2.5pt}{0.95}
\small
\centering
\begin{tabular}{lcllcllclll}
\shline
\multirow{3}{*}{Method} && \multicolumn{5}{c}{CIFAR10-LT: \# of labels (percentage)} && \multicolumn{3}{c}{CIFAR10: \# of labels (percentage)} \\
\cline{3-7}\cline{9-11}
\multirow{1}{*}{} && \multicolumn{2}{c}{$\gamma$=100} && \multicolumn{2}{c}{$\gamma$=200} && \multirow{2}{*}{40 (0.08\%)} & \multirow{2}{*}{80 (0.16\%)} & \multirow{2}{*}{250 (2\%)} \\
\cline{3-4}\cline{6-7}
\multirow{1}{*}{} && \multicolumn{1}{c}{1244 (10\%)} & \multicolumn{1}{c}{3726 (30\%)} && \multicolumn{1}{c}{1125 (10\%)} & \multicolumn{1}{c}{3365 (30\%)} && \multicolumn{1}{c}{} & \multicolumn{1}{c}{} & \multicolumn{1}{c}{} \\
\cline{1-1}\cline{3-4}\cline{6-7}\cline{9-11}
\cline{1-1}\cline{3-4}\cline{6-7}\cline{9-11}
\cline{1-1}\cline{3-4}\cline{6-7}\cline{9-11}
UDA \cite{xie2019unsupervised} $\S$ && - & - && - & - && 71.0 $\pm$6.0 & - & 91.2 $\pm$1.1  \\
MixMatch \cite{berthelot2019mixmatch} $\S$ && 60.4 $\pm$2.2 & - && 54.5 $\pm$1.9 & - && 51.9 $\pm$11.8 & 80.8 $\pm$1.3 & 89.0 $\pm$0.9 \\
CReST w/ DA \cite{wei2021crest}  && 75.9 $\pm$0.6 & 77.6 $\pm$0.9 && 64.1 $\pm$0.22 & 67.7 $\pm$0.8 && - & - & - \\
CReST+ w/ DA \cite{wei2021crest}  && 78.1 $\pm$0.8 & 79.2 $\pm$0.2 && 67.7 $\pm$1.4 & 70.5 $\pm$0.6 && - & - & - \\
CoMatch w/ SimCLR \cite{li2021comatch,chen2020simple} && - & - && - & - && 92.6 $\pm$1.0 & 94.0 $\pm$0.3 & 95.1 $\pm$0.3 \\
\cline{1-1}\cline{3-4}\cline{6-7}\cline{9-11}
FixMatch \cite{sohn2020fixmatch} $\S$ && 67.3 $\pm$1.2 & 73.1 $\pm$0.6 && 59.7 $\pm$0.6 & 67.7 $\pm$0.8 && 86.1 $\pm$3.5 & 92.1 $\pm$0.9 & 94.9 $\pm$0.7 \\
FixMatch w/ DA w/ LA \cite{wei2021crest,sohn2020fixmatch,berthelot2019remixmatch,menon2021long} $\S$ && 70.4 $\pm$2.9 & - && 62.4 $\pm$1.2 & - && - & - & - \\
FixMatch w/ DA w/ SimCLR \cite{sohn2020fixmatch,berthelot2019remixmatch,chen2020simple} $\S$ && - & - && - & - && 89.7 $\pm$4.6 & 93.3 $\pm$0.5 & 94.9 $\pm$0.7 \\
\hline
DebiasPL (w/ FixMatch) &&\bf 79.2 $\pm$1.0 & \bf 80.6 $\pm$0.5 && \bf 71.4 $\pm$2.0 & \bf 74.1 $\pm$0.6 && \bf 94.6 $\pm$1.3 & \bf 95.2 $\pm$0.1 & \bf 95.4 $\pm$0.1 \\
\it gains over the best FixMatch variant && \color{green(pigment)} \bf +8.8 & \color{green(pigment)} \bf +7.5 && \color{green(pigment)} \bf +9.0 & \color{green(pigment)} \bf +6.4 &&\color{green(pigment)} \bf +4.9 & \color{green(pigment)} \bf +1.9 & \color{green(pigment)} \bf +0.5 \\
\shline
\end{tabular}\vspace{-4pt}
\caption{Without any prior knowledge of the marginal class distribution of unlabeled/labeled data, the performance of DebiasPL on \textit{both} \textbf{CIFAR \textit{and} CIFAR-LT SSL benchmarks} surpasses previous SOTAs, which are \textit{either designed for balanced data or meticulously tuned for long-tailed data}.
DibasMatch is experimented with the same set of hyper-parameters across all benchmarks.
$\S$ states the best-reported results of counterpart methods, copied from \cite{li2021comatch}, \cite{sohn2020fixmatch} or \cite{wei2021crest}.
$\gamma$: imbalance ratio.
We report results averaged on 5 different folds.
}
\label{table:cifar}
\end{table*}
}

\def\tabImageNet#1{
\begin{table*}[#1]
\tablestyle{3.5pt}{0.92}
\small
\centering
\begin{tabular}{lcccccccllcll}
\shline
\multirow{2}{*}{Method} && \multirow{2}{*}{B.S.} && \multirow{2}{*}{\#epochs} && \multirow{2}{*}{Pre-train} && \multicolumn{2}{c}{1\%} && \multicolumn{2}{c}{0.2\%} \\
\cline{9-10}\cline{12-13}
\multirow{1}{*}{} && \multirow{1}{*}{} && \multirow{1}{*}{} && \multirow{1}{*}{} && \multicolumn{1}{c}{top-1} & \multicolumn{1}{c}{top-5} && \multicolumn{1}{c}{top-1} & \multicolumn{1}{c}{top-5} \\
\cline{1-1}\cline{3-3}\cline{5-5}\cline{7-7}\cline{9-10}\cline{12-13}
\cline{1-1}\cline{3-3}\cline{5-5}\cline{7-7}\cline{9-10}\cline{12-13}
FixMatch w/ DA \cite{sohn2020fixmatch, berthelot2019remixmatch} && 4096 && 400 && $\xmark$ && 53.4 & 74.4 && - & - \\
FixMatch w/ DA \cite{sohn2020fixmatch, berthelot2019remixmatch} && 4096 && 400 && $\cmark$ && 59.9 & 79.8 && - & - \\
FixMatch w/ EMAN \cite{sohn2020fixmatch,cai2021exponential} && 384 && 50 && $\cmark$ && 60.9 & 82.5 && 43.6$^*$ & 64.6$^*$ \\
\hline
DebiasPL w/ FixMatch && 384 && 50 && $\cmark$ && \bf 63.1 \small\color{green(pigment)} (+2.2) & \bf 83.6 \small\color{green(pigment)} (+1.1) && \bf 47.9 \small\color{green(pigment)} (+3.7) & \bf 69.6 \small\color{green(pigment)} (+5.0)\\
DebiasPL (multi-views) && 768 && 50 && $\cmark$ && \bf 65.3 \small\color{green(pigment)} (+4.4) & \bf 85.2 \small\color{green(pigment)} (+2.7) && \bf 51.6 \small\color{green(pigment)} (+8.0) & \bf 73.3 \small\color{green(pigment)} (+8.7) \\
DebiasPL (multi-views) && 768 && 200 && $\cmark$ && \bf 66.5 \small\color{green(pigment)} (+5.6) & \bf 85.6 \small\color{green(pigment)} (+3.1) && \bf 52.3 \small\color{green(pigment)} (+8.7) & \bf 73.5 \small\color{green(pigment)} (+8.9) \\
DebiasPL (multi-views) && 1536 && 300 && $\cmark$ && \bf 67.1 \small\color{green(pigment)} (+6.2) & \bf 85.8 \small\color{green(pigment)} (+3.3) && - & - \\
DebiasPL w/ CLIP \cite{radford2021learning} && 384 && 50 && $\cmark$ && \bf 69.1 \small\color{green(pigment)} (+8.2) & \bf 89.1 \small\color{green(pigment)} (+6.6) && \bf 68.2 \small\color{green(pigment)} (+24.6) & \bf 88.2 \small\color{green(pigment)} (+23.6) \\
DebiasPL w/ CLIP (multi-views) \cite{radford2021learning} && 768 && 50 && $\cmark$ && \bf 70.9 \small\color{green(pigment)} (+10.0) & \bf 89.3 \small\color{green(pigment)} (+6.8) && \bf 69.6 \small\color{green(pigment)} (+26.0) & \bf 88.4 \small\color{green(pigment)} (+23.8) \\
\hline 
CLIP (few-shot) \cite{radford2021learning,zhou2021learning} && 256 && 50 && $\cmark$ && 53.4 & - && 40.0 & - \\
SwAV \cite{caron2020unsupervised} && 4096 && 50 && $\cmark$ && 53.9 & 78.5 && - & - \\
SimCLRv2 (+ Self-distillation) \cite{chen2020big} && 4096 && 400 && $\cmark$ && 60.0 & 79.8 && - & - \\
PAWS (multi-crops) $\dagger$ \cite{assran2021semi} && 4096 && 50 && $\cmark$ && 66.5 & \color{gray}- && - & - \\
CoMatch (multi-views) \cite{li2021comatch} && 1440 && 400 && $\cmark$ && 67.1 & 87.1 && - & - \\
\cline{1-1}\cline{3-3}\cline{5-5}\cline{7-7}\cline{9-10}\cline{12-13}
\shline
\end{tabular}\vspace{-4pt}
\caption{DebiasPL delivers state-of-the-arts results on
\textbf{ImageNet-1K semi-supervised learning}  with various fractions of labeling samples, especially for extremely low-shot settings.
All results are produced with a backbone of ResNet-50. $\dagger$: unsupervised pre-trained for 800 epochs, except for PAWS \cite{assran2021semi}, which is pre-trained for 300 epochs with pseudo-labels generated non-parametrically.
$^*$: reproduced. 
}
\label{table:imagenet}
\end{table*}
}

\def\tabfigure#1#2#3{
    \begin{subfigure}{#2\linewidth}
        \includegraphics[width=1.0\linewidth]{#1}\vspace{-2pt}
        \caption{#3}
    \end{subfigure}
}

\def\figBiasCLIP#1{
    \begin{figure}[#1]
    \centering
    \tb{@{}ccc@{}}{0}{
        \tabfigure{figures/teaser_clip-f.pdf}{0.3}{ImageNet} & 
        \tabfigure{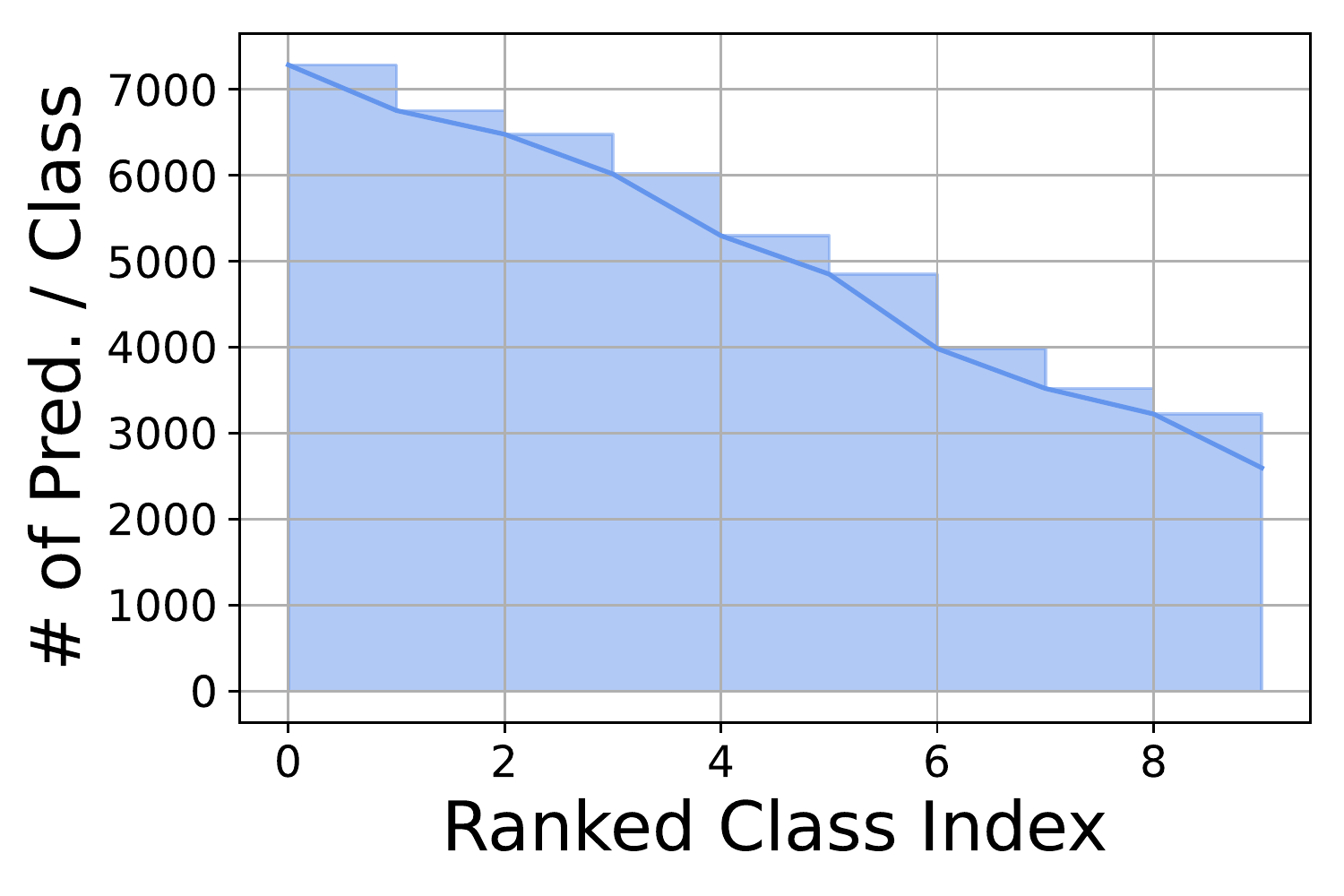}{0.3}{CIFAR10} & 
        \tabfigure{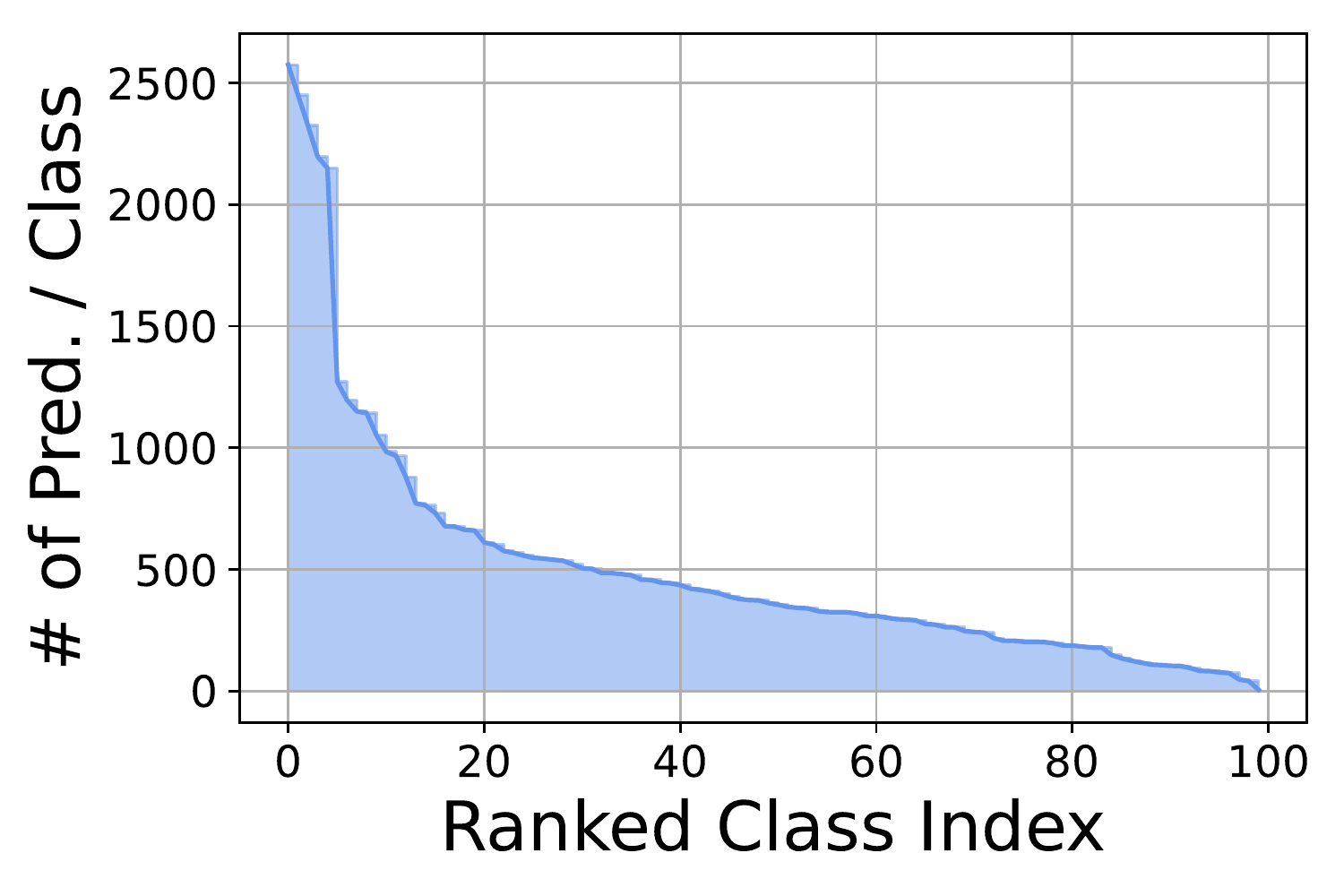}{0.3}{CIFAR100} \\
        \tabfigure{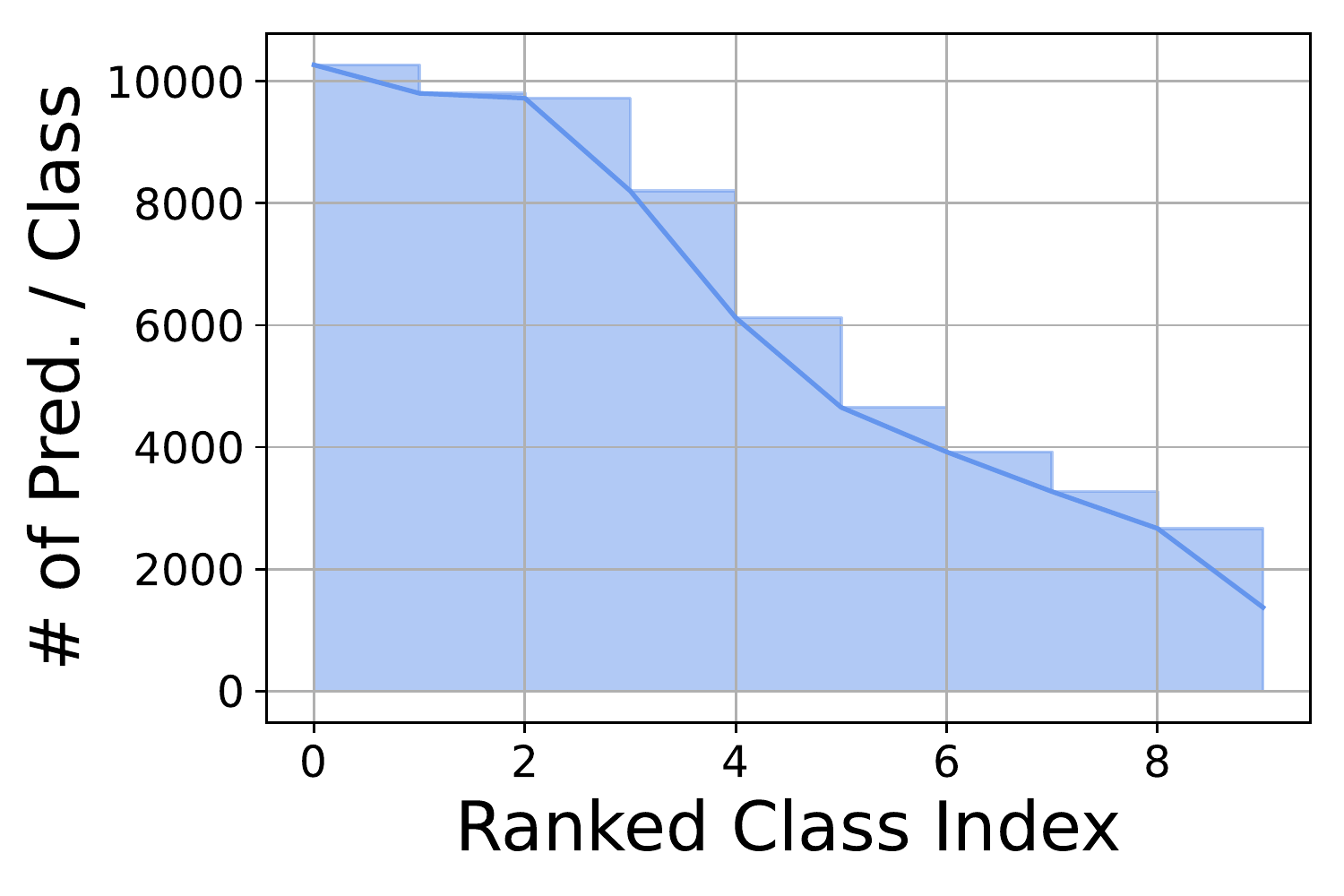}{0.3}{MNIST} & 
        \tabfigure{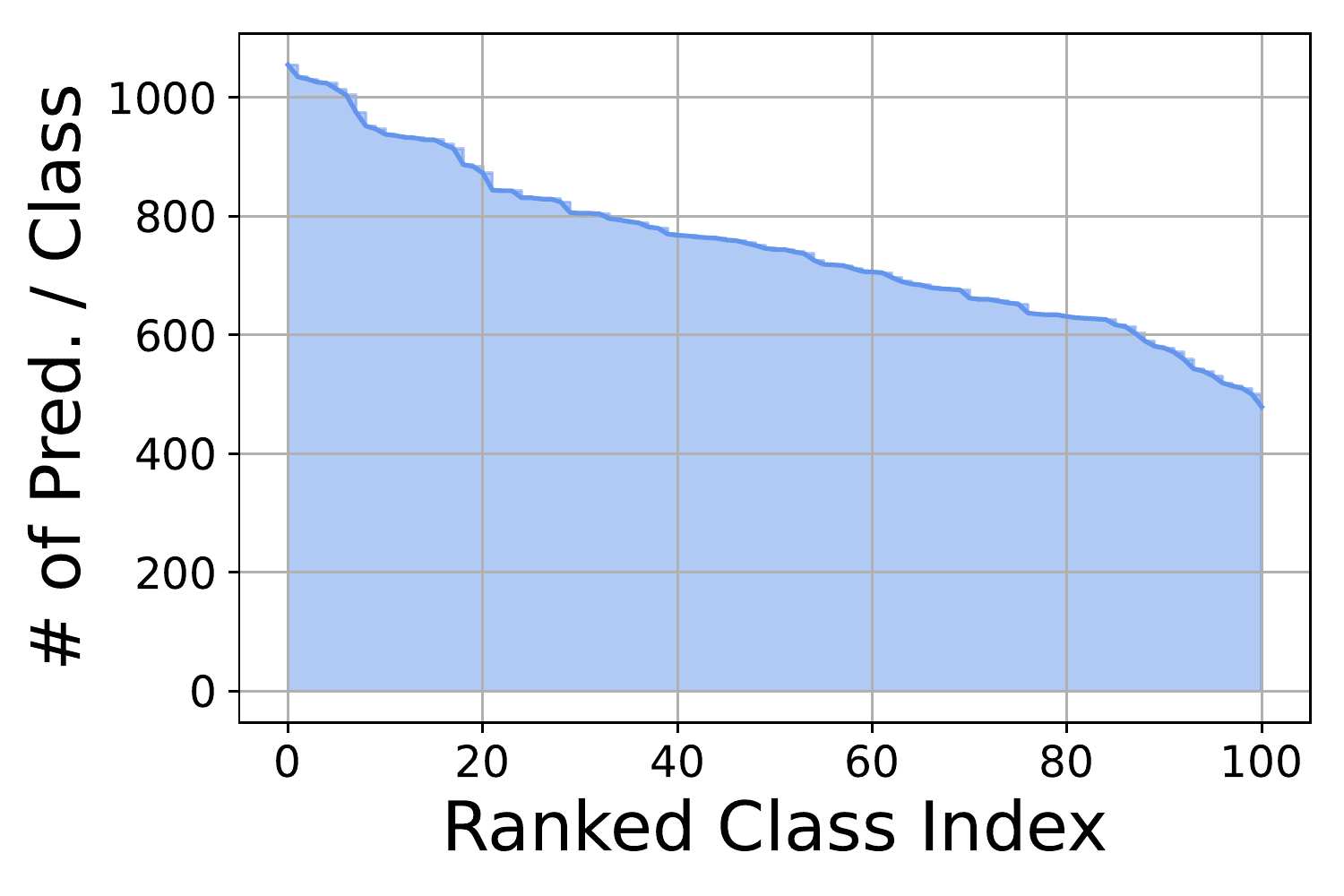}{0.3}{Food101} & 
        \tabfigure{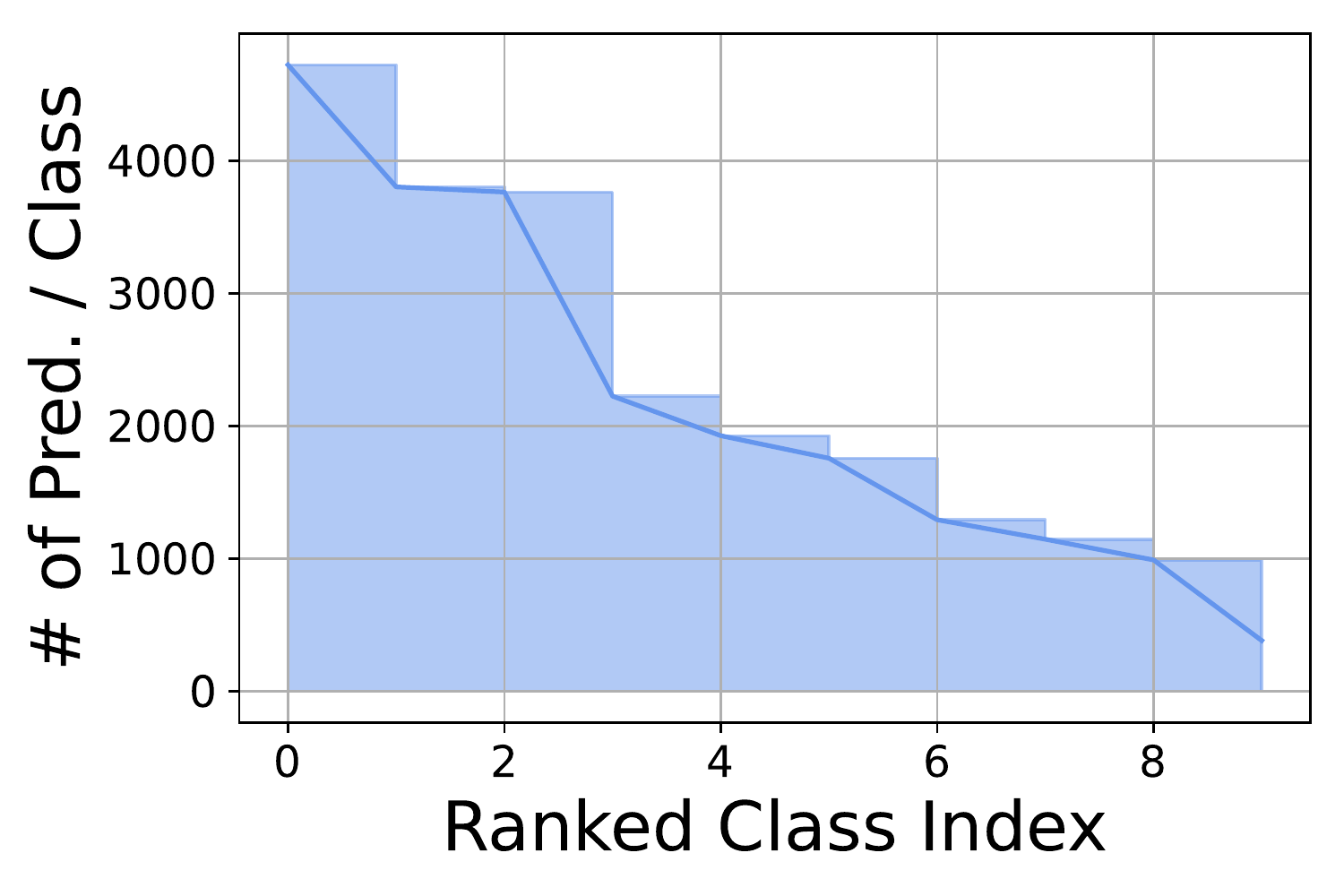}{0.3}{EuroSAT}
    }
    \vspace{-6pt}
    \caption{
        {CLIP’s zero-shot predictions are highly biased for various datasets and benchmarks.} 
    }
    \label{fig:bias-clip-datasets}
    \end{figure}
}

\section{Pseudo-Labels are Naturally Imbalanced}
\label{bias}
In contrast to previous work that concentrated on biases caused by trained on \textit{imbalanced} data, our focus is on pseudo-label biases, even when trained on \textit{balanced} data.
In this section, we provide an analysis of this previously unknown issue hidden behind the tremendous success of FixMatch \cite{sohn2020fixmatch} on SSL and CLIP \cite{radford2021learning} on ZSL, both of which require the use of  ``pseudo-labeling" to transfer knowledge learned in source data to target data. 

We first describe the backgrounds for pseudo-labeling approaches and then analyze their bias issue. We attribute the cause of bias to the inter-class correlation problem.

\figBiasFixMatch{t}

\subsection{Background}
\noindent\textbf{FixMatch for semi-supervised learning.} The core technique of FixMatch \cite{sohn2020fixmatch} is pseudo-labeling \cite{lee2013pseudo}. It selects unlabeled samples with high confidence as training targets. 

Suppose we have a labeled dataset $X_L = \{(x_i, y_i)\}_{i=1}^{L}$ with $L$ labeled instances, and an unlabeled dataset $X_U = \{(x_i)\}_{i=L+1}^{L+U}$ with $U$ instances. $x_i$ is the input instance and $y_i = [y_i^1,...,y_i^C] \subseteq \{0, 1\}^C$ is a discrete annotated target with $C$ classes. $X_U$ and $X_L$ share the same semantic labels.
The optimization objective consists of two terms: $\mathcal{L} = \mathcal{L}_{\text{s}} + \lambda_u\mathcal{L}_{\text{u}}$, i.e., the supervised loss $\mathcal{L}_s$ applied to labeled data and an unsupervised loss $\mathcal{L}_u$ applied to unlabeled data, and $\lambda_u$ is a scalar hyperparameter.

The supervised loss $\mathcal{L}_s$ is the cross-entropy between the model predictions and the ground truth: $\mathcal{L}_{\text{s}} = \frac{1}{B}\sum_{i=1}^B\text{H}(y_i, p(\alpha(x_i)))$, where $\alpha$ is the weak augmentation, and $B$ is the batch size.
 The pseudo-labels $\hat{y_i}$ for unlabeled instances are generated from the weakly-augmented unlabeled samples, which are used to supervise the model prediction of the strongly-augmented samples.
Instances whose largest probability fall under a confidence threshold $\tau$ are regarded as unreliable samples and discarded.
Formally, the unsupervised loss $\mathcal{L}_u$ can be formulated as:
\begin{equation}
    \!\!\mathcal{L}_{\text{u}} = \frac{1}{\mu B}\sum_{i=1}^{\mu B}\mathbbm{1}[\max ( p(\alpha(x_i)) ) \ge \tau] \cdot \text{H}(\hat{y_i}, p(\beta(x_i)))
    \label{fixmatch-uloss}
\end{equation}
\noindent where $\beta$ is a strong augmentation \cite{cubuk2020randaugment}, and $\mu$ determines the ratio of labeled and unlabeled samples in the minibatch.

\noindent\textbf{CLIP for zero-shot learning.} 
CLIP \cite{radford2021learning} is an efficient and scalable way to learn image representations from scratch on a dataset of 400M image-text pairs, which is manually curated to be approximately query-balanced.
At pre-training time, an image encoder and a text encoder are optimized by maximizing (minimizing) the similarity between paired (unpaired) captions and visual images.

For producing pseudo-labels of unlabeled data, natural language prompting is used to enable zero-shot transfer to target datasets: CLIP uses the names or descriptions of the target dataset’s classes as the set of potential text pairings (e.g. ``a photo of a dog") and predicts the most probable class according to the cosine similarity of image-text pairs. Specifically, the feature embedding of the image and the feature embedding of the set of possible texts are first computed by their respective encoders. The cosine similarity of these embeddings is then evaluated, and normalized into a probability distribution via a softmax function.

\subsection{Biases in Semi-supervised Learning}

\figCLIPBias{t}
\figBiasCLIP{t}

\fig{bias-fixmatch} visualizes the FixMatch probability distributions averaged on all unlabeled data at various training epochs.
Surprisingly, even when labeled and unlabeled data are both curated (class-balanced), the pseudo-labels are still highly class-imbalanced, most notably at the early training stage. 
As the training progresses, this situation persists.

\figInterClassCorrCLIP{ht}

A student model will inherit the implicitly imbalanced pseudo-labels and, in turn, reinforces the teacher model's biases. Once confusing samples are wrongly pseudo-labeled, the mistake is almost impossible to be self-corrected. On the contrary, it may even mislead the model and further amplify existing bias to produce more wrong predictions. Without intervention, the model will get trapped in irreparable biases. 

On the contrary, as in \fig{bias-fixmatch}, although DebiasPL is also troubled by the imbalanced pseudo-labels at the beginning, this situation can be significantly alleviated, and, eventually, we can obtain an almost balanced distribution through dynamically debiasing the model.

\subsection{Biases in Zero-Shot Learning}
CLIP actually generates highly biased predictions on ImageNet, which is hidden behind CLIP's tremendous success in terms of overall zero-shot prediction accuracy.

Except for the imbalance problem, the precision and recall of many high-frequency classes are much lower than many medium-/few-shot classes, as illustrated in Fig.~\ref{fig:clip-bias}.
Thresholding the CLIP predictions based on the confidence score may help.
However, simply setting a higher confidence score threshold could lead to even more imbalanced distributions (more details in appendix). There is a trade-off between imbalance ratio and precision/recall.

\figInterClassCorrFixMatch{t}

Highly biased zero-shot predictions are not unique to ImageNet. They are widely present on many benchmarks, such as EuroSAT \cite{helber2019eurosat}, MNIST \cite{lecun1998mnist}, CIFAR10 \cite{krizhevsky2009learning}, CIFAR100 \cite{krizhevsky2009learning}, and Food101 \cite{bossard14}, as shown in \fig{bias-clip-datasets}.

\subsection{Inter-Class Correlations} 
To delve into the causes of biased pseudo-labels, we provide an analysis of inter-class correlations.
For CLIP, we first compute one image centroid per class by taking the mean of the normalized image features, extracted by the image encoder of a pre-trained CLIP model, that belong to this class. The cosine similarity between the image centroid of classes with top-10/least-10 prediction frequency and their closest ``confusing" classes are visualized. 
The prediction confusions indicate image similarities at the class level. Fig.~\ref{fig:inter-class-clip} shows that the low-frequency classes of ImageNet, with the least-10 number of CLIP predictions per class, usually have strong inter-class confusions.

Fig.~\ref{fig:inter-class-fixmatch-a} shows the confusion matrix of FixMatch's pseudo-labels. It is observed that many instances in some categories tend to be misclassified into one or two specific negative classes; for instance, ``ship" is often misclassified as ``plane".

Based on our analysis of the inter-class correlations, we believe that the blame for the pseudo-label bias can be largely attributed to inter-class confounding, which the proposed DebiasPL can successfully address as in Fig.~\ref{fig:inter-class-fixmatch-b}. DebiasPL will be introduced in the next section.

\section{Debiased Pseudo-Labeling}

This section introduces Debiased Pseudo-Labeling (DebiasPL) and methods to integrate it into ZSL and SSL tasks. 
It is worth noting that the proposed simple yet effective approach is universally applicable to various networks and benchmarks, not limited to the ones introduced here.

\subsection{Adaptive Debiasing}
\label{adaptive-debias}
Our DebiasPL approach aims at dynamically alleviating biased pseudo labels' influence on a student model without leveraging any prior knowledge on marginal class distribution, even when exposed to source and target data that follow different distributions.
An adaptive debiasing module with counterfactual reasoning and an adaptive marginal loss is proposed to fulfill this goal, described next.

\figdDiagramDebias{ht}
\figdCausal{ht}

\noindent\textbf{Adaptive Debias w/ Counterfactual Reasoning}.
{Causal Inference} is the undertaking of deriving counterfactual conclusions using only factual premises, in which causal graphical models represent the interventions among the variables \cite{pearl2013direct,pearl2009causal,greenland1999confounding,rubin2019essential,rubin2005causal}. 
It has been widely studied and applied in various tasks to remove selection bias which is pervasive in almost all empirical studies \cite{bareinboim2012controlling}, eliminating the confounding effect using causal intervention \cite{zhang2020causal}, disentangling the desired direct effects with counterfactual reasoning \cite{besserve2019counterfactuals}, etc. 

Motivated by this, to dynamically mitigate impacts of unwanted bias (\textit{counterfactual}), we incorporate causality of producing debiased predictions through counterfactual reasoning \cite{holland1986statistics, pearl2009causal, pearl2009causality, pearl2013direct, pearl2018book}.

Given the proposed causal graph in \fig{causal-graph}, we can delineate our goal for generating debiased predictions: the pursuit of the direct causal effect along $A_i \rightarrow Y$, defined as Controlled Direct Effect (CDE) \cite{pearl2013direct,richiardi2013mediation,pearl2018book,greenland1999confounding,tang2020long}:
\begin{equation}
    \text{CDE}(Y_{i}) = [Y_{i}|do(A_i), do(D)] - [Y_{i}|do(\hat{A}), do(D)]
\end{equation}
i.e. the contrast between the counterfactual outcome if the individual were exposed at $A=A_i$ (with $do(A_i)$ notation) and the counterfactual outcome if the same individual were exposed at $A=\hat{A}=\{A_1,...,A_n\}$, with the mediator set to a fixed level $D$.
CDE \cite{pearl2013direct,greenland1999confounding} disentangles the model bias in a counterfactual world, where the model bias is considered as the $Y$’s indirect effect when $A=\hat{A}$ but $D$ retains the value when $A=A_i$.

However, measuring the counterfactual outcome via visiting all training samples is significantly computational expensive. We use Approximated Controlled Direct Effect (ACDE) instead. ACDE assumes that the model bias is not drastically changed, therefore, the momentum-updated counterfactual outcomes (Eqn.~\ref{momentum}) can be served as an approximation to the actual $[Y_{i}|do(\hat{A}), do(D)]$. The debiased logits with counterfactual reasoning, which is later used to perform pseudo-labeling (i.e., replace $p(\alpha(x_i))$ in Eqn.~\ref{fixmatch-uloss}), can be formulated:
\begin{align}
&\tilde{f_i} = f(\alpha(x_i)) - \lambda\log\hat{p} \label{acde} \\
&\hat{p} \leftarrow m\hat{p} + (1-m)\frac{1}{\mu B}\sum_{k=1}^{\mu B}p_{k}
\label{momentum}
\end{align}
\noindent $m\in[0, 1)$ is a momentum coefficient,
$f(\alpha(\cdot))$ refers to logits of weakly-augmented unlabeled instance, 
$p_{k}$ is the probability distribution for instance $\alpha(x_k)$ obtained via a softmax function.
$\lambda$ denotes the debias factor, which controls the strength of the indirect effect. If the debias factor is too strong, it is hard for a model to fit on the data, while too small a factor can barely eliminate the biases and, ultimately, impairs the generalization ability. Since the scale of logits is unstable, most notably at the early training stage, we use the probability distribution $p_k$ rather than directly using the logit vector in the second term of Eqn.~\ref{acde}. 
A log function is applied to rescale $\hat{p}$ to match the magnitude of logit. 

Eqn.~\ref{acde} can be associated with re-weighting and logits adjustment methods in long-tailed recognition, whereas ours is dynamically adaptive. 

\noindent\textbf{Adaptive Marginal Loss.}
As aforementioned in Sec.~\ref{bias}, the biases in pseudo-labels may be partially caused by inter-class confusion. Motivated by this, we apply adaptive margin loss to demand a larger margin between hardly biased and highly biased classes, so that scores for dominant classes, towards which the model highly biased, do not overwhelm the other categories. In addition, by enforcing a dynamic class-specific margin, inter-class confusion can be greatly counteracted, which is further empirically evidenced in Fig.~\ref{fig:inter-class-fixmatch}. $\mathcal{L}_{\text{AML}}$ can be formulated as:
\begin{equation}
\mathcal{L}_{\text{AML}} = -\textrm{log}\frac{e^{(z_{\hat{y_i}} - \Delta_{\hat{y_i}})}}{e^{(z_{\hat{y_i}} - \Delta_{\hat{y_i}})} + \sum_{k\neq \hat{y_i}}^{C}e^{(z_{k}-\Delta_k)}}
\label{aml}
\end{equation}
\noindent where $\Delta_j=\lambda\log(\frac{1}{\hat{p}_j}) \text{ for } j\in\{1,...,C\}$, $z=f(\beta(x_i))$. We use $\mathcal{L}_{\text{AML}}$ to replaced $\text{H}(\hat{y_i}, f(\beta(x_i))$ in Eqn.~\ref{fixmatch-uloss}.
We then get the final unsupervised loss by updating Eqn.~\ref{fixmatch-uloss} with Eqn.~\ref{acde} and Eqn.~\ref{aml}.

\noindent (Optional) All unlabeled instances with low probabilities do not contribute to the final loss. We find it beneficial to apply cross-level instance-group discrimination loss CLD \cite{wang2021unsupervised} to unlabeled instances to leverage their information fully. 

\subsection{Distinctions and Connections with Alternatives} 
\label{connection-distinctions}
Please refer to Sec.~\ref{related-work-long-tailed} for an introduction to LA, LDAM, and Causal Norm. Another often adopted method in SSL distribution alignment (DA) \cite{berthelot2019remixmatch} is also compared. It aims to encourage the actual marginal distribution of the model’s predictions to match the \textit{actual} marginal class distribution.

Please refer to Tab.~\ref{tab:distMethods} to check the distinctions and connections with these alternatives handling distribution mismatch and long-tailed recognition in key properties, and Tab.~\ref{table:cifar} and Tab.~\ref{table:imagenet} to compare experimental results. 

The use of a momentum updated $\hat{p}$ for debiasing pseudo-labels with counterfactual reasoning and applying adaptive marginal loss is crucial to the success of DebiasPL, which also enables our training objective does not necessarily need to use the true marginal class distribution as prior knowledge. 
Furthermore, since more training samples per class \textit{do not} necessarily lead to a higher model bias against it, dynamically adjusting the margin rather than measuring margins based on the number of samples per class as in LA and LDAM could better respect the degree of bias against each class. The number of samples alone can not determine the degree of bias. 
Also, unlike previous works, e.g., LA/LDAM and Causal Norm, that use fixed margins or adjustments, we argue that the degree of bias of each class should never be a fixed value, but is in a process of dynamic change. The cause of bias cannot be attributed to the data alone, but the cause of the interaction between model and data.

For DA, the biggest issue is that it is limited to scenarios where either \textit{true} marginal class distribution is available, or source and target data are collected from the same distribution, which is too ideal in the real world. 

Experiments on several benchmarks are made to show the validity and feasibility of DebiasPL. 
\textit{For imbalanced data}, Tab.~\ref{tab:distMethods} shows that integrating LA \cite{menon2021long} into FixMatch lags far behind FixMatch w/ DebiasPL.
\textit{For balanced data}, since the adjustment or re-weighting vector is calculated based on the true class distribution, most existing long-tailed methods that rely on true marginal class distribution are no longer applicable without major changes (balanced class distribution leads to identical treatment for all classes).

\tabDistMethods{t}

\tabCIFAR{ht}

\tabImageNet{ht}

\subsection{DebiasPL for T-ZSL and SSL}
\noindent\textbf{For semi-supervised learning}, the proposed DebiasPL can be integrated into FixMatch, as in \fig{Diagram}, by adopting the adaptive debiasing module and adaptive marginal loss. To further boost the performance of SSL and exploit the power of the vision-language pre-trained model, during the training time, we can also integrate CLIP into FixMatch/DebiasPL by pseudo-labeling the discarded unlabeled instances with CLIP. 
Because the instances CLIP are not confident on may be noisy, only these unlabeled instances with a CLIP confidence score greater than $\tau_{\text{clip}}$ are pseudo-labeled by CLIP.
We could get CLIP's predictions on all training data and store it in a dictionary without re-predicting per iteration. Therefore, the computational overheads introduced by using the CLIP model are negligible.
We only leverage CLIP in large-scale datasets since using CLIP on low-resolution datasets like CIFAR10 can only observe marginal gains, partly due to the lack of scale-based data augmentation in CLIP \cite{radford2021learning}.

\noindent\textbf{For transductive zero-shot learning}, to better exploit knowledge learned from the vision-language pre-trained model and alleviate the domain shift problem when transferring the knowledge to downstream ZSL tasks, a new framework to conduct transductive zero-shot learning (T-ZSL) based on FixMatch and CLIP is developed.

Specifically, we again make use of the \textit{pseudo-labeling} idea by leveraging the one-hot labels (i.e., the $\arg\max$ of the model’s output) and retaining pseudo labels whose largest class probability fall above a confidence threshold $\tau_{\text{clip}}$ ($=0.95$ by default). These instances, along with their pseudo labels, are considered ``labeled data" in SSL. 

After this, we could  follow the original FixMatch pipeline to  optimize ``labeled" and ``unlabeled" data jointly.
To make a fair comparison with previous works and simplify the overall system, all other training recipes and settings are consistent with the original FixMatch+EMAN settings, including the model initialization part. The diagram is in the appendix.

Because CLIP is highly biased, the vanilla FixMatch + CLIP framework under-performs the original CLIP zero-shot learning, confirming our earlier hypothesis that learning from a biased model may further amplify existing bias and produce more wrong predictions. Therefore, we update the unsupervised loss $\mathcal{L}_u$ with our Adaptive Marginal Loss for alleviating the inter-class confusion and Adaptive Debias for producing debiased pseudo-labels as in Sec.~\ref{adaptive-debias}. 
\def\tabImageNetZeroShot#1{
\begin{table}[#1]
\tablestyle{2pt}{0.97}
\small
\centering
\begin{tabular}{l||l|ll}
\shline
\multirow{2}{*}{Method} & \multirow{2}{*}{\#param} & \multicolumn{2}{c}{Accuracy (\%)} \\ 
\multirow{1}{*}{} & \multirow{1}{*}{} & \multicolumn{1}{c}{top-1} & \multicolumn{1}{c}{top-5} \\
\shline
ConSE \cite{norouzi2014zero} & - & 1.3 & 3.8 \\
DGP  \cite{kampffmeyer2019rethinking} & - & 3.0 & 9.3 \\
ZSL-KG \cite{nayak2020zero} & - & 3.0 & 9.9 \\
Visual N-Grams \cite{li2017learning} & - & 11.5 & - \\
\hline 
CLIP (prompt ensemble) \cite{radford2021learning} & 26M & 59.6 & - \\
(ours) CLIP + FixMatch & 26M & 55.7 & 80.6 \\
(ours) CLIP + DebiasPL & 26M & \bf 68.3 \small\color{green(pigment)} (+8.7) & \bf 88.9 \small\color{green(pigment)} (+8.3) \\
\hline
\color{gray} CLIP (few-shot) \cite{radford2021learning,zhou2021learning} $\dagger$ &\color{gray} 26M & \color{gray} 53.4 & - \\
\color{gray} CLIP + CoOp (few-shot) \cite{zhou2021learning} $\dagger$ &\color{gray} 26M &\color{gray} 60.9 & - \\
\hline 
\color{gray} CLIP (ViT-B/32) \cite{radford2021learning} & \color{gray} 398M & \color{gray} 63.2 & - \\
\color{gray} CLIP (ResNet50x4) \cite{radford2021learning} & \color{gray} 375M & \color{gray} 65.8 & - \\
\shline
\end{tabular}\vspace{-4pt}
\caption{DebiasPL delivers state-of-the-art results of \textbf{zero-shot learning on ImageNet-1K}, outperforming CLIP with bigger models or fine-tuned with labels. 
$\dagger$: CoOp and CLIP (few-shot) are fine-tuned with about 1.5\% annotated data.
}
\label{table:zeroshot}
\end{table}
}

\def\tabAblationSSLMethods#1{
\begin{table}[#1]
\tablestyle{7pt}{0.97}
\small
\centering
\begin{tabular}{l||cccccc}
\shline
 & FixMatch & MixMatch & UDA \\ 
\shline
Baseline & 89.7 $\pm$ 4.6  & 47.5 $\pm$ 11.5 & 29.1 $\pm$ 5.9 \\
+ DebiasPL & \bf 94.6 $\pm$ 1.3 & \bf 61.7 $\pm$ 6.1 & \bf 43.2 $\pm$ 5.2 \\
\shline
\end{tabular}\vspace{-4pt}
\caption{\textbf{DebiasPL is a universal add-on.} Top-1 accuracies of various SSL methods on CIFAR10, averaged on 5 folds, are compared. 4 instances per class are labeled.
}
\label{table:methods-SSL}
\end{table}
}

\def\figAblationZeroShotDatasets#1{
    \captionsetup[sub]{font=small}
    \begin{figure}[#1]
        \centering
        \includegraphics[width=0.96\linewidth]{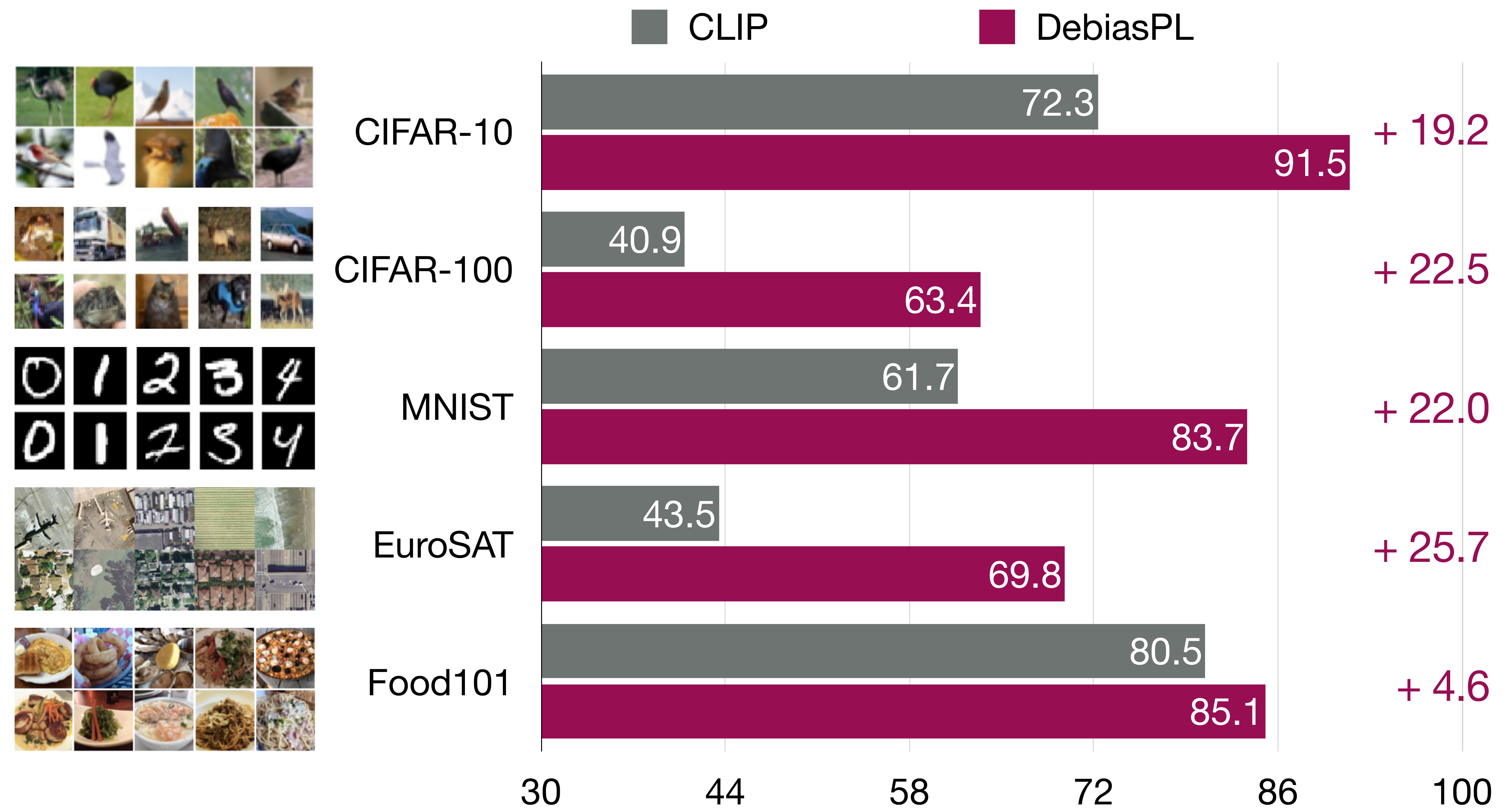}
        \vspace{-4pt}
        \caption{DebiasPL exhibits stronger robustness to domain shift when conducting \textbf{zero-shot learning on various datasets}. We experiment with ResNet-50 as a backbone network. CLIP results are reproduced with official codes.}
        \label{fig:datasets-ZSL}
    \end{figure}
}

\def\tabCIFARDist#1{
\begin{table}[#1]
\tablestyle{7pt}{0.97}
\small
\centering
\begin{tabular}{l||l|l}
\shline
\multirow{2}{*}{Method} & \multicolumn{2}{c}{Labeled: \textbf{LT}; 10\% labeled, $\gamma=200$} \\ 
\cline{2-3}
\multirow{1}{*}{} & \multicolumn{1}{c|}{Unlabeled: \textbf{LT}} & \multicolumn{1}{c}{Unlabeled: \textbf{Balanced}} \\ 
\shline
FixMatch \cite{sohn2020fixmatch} & 62.3 $\pm$1.6 & 72.1 $\pm$2.3 \\
DebiasPL & \bf 71.4 $\pm$2.0 \small\color{green(pigment)} (+9.1) & \bf 83.5 $\pm$2.4 \small\color{green(pigment)} (+11.4) \\
\shline
\end{tabular}\vspace{-4pt}
\caption{DebiasPL consistently improves the performance of SSL when the unlabeled data is either the sames as labeled data, i.e., long-tailed distributed, or different with labeled data, i.e., balanced distributed across semantics. 
We report results averaged on 5 folds.
}
\label{table:cifar-dist}
\end{table}
}

\section{Experiment}
In this section, we conduct empirical experiments to show that DebiasPL: 1) delivers state-of-the-art results on both semi-supervised and zero-shot learning benchmarks; 2) works as a universal add-on and brings consistent performance gains to various methods; 3) exhibits stronger robustness to domain shifts; 4) is capable of improving performance on long-tailed, balanced and even hybrid data.

\subsection{Semi-supervised Learning}
\noindent\textbf{Dataset}. We perform comprehensive evaluations of DebiasPL on multiple SSL benchmarks, including CIFAR10 \cite{krizhevsky2009learning}, long-tailed CIFAR10 (CIFAR10-LT) \cite{krizhevsky2009learning}, and ImageNet-1K \cite{ILSVRC15}, with varying amounts of labeled data. 
For the balanced benchmarks, the performance almost saturates when using more than 2\% labeled data. We put our focus on the extremely low-shot settings, i.e., 0.08\%/0.16\%/2\% on CIFAR10 and 1\%/0.2\% on ImageNet-1K. For imbalanced benchmarks, we follow the settings in \cite{wei2021crest} and test DebiasPL on CIFAR10-LT under various pre-defined imbalance ratios $\gamma$, where $\gamma\in[100,200]$, and percentage of labeled data, including 10\% and 30\%. More details about datasets are included in the appendix.

\noindent\textbf{Setup}.
For all experiments on \textit{both} long-tailed CIFAR10 and CIFAR10 datasets, we follow previous works \cite{sohn2020fixmatch, wei2021crest} to use the network architecture  WRN-28-2 \cite{he2016deep, zagoruyko2016wide}. 
We also follow the same set of hyper-parameters in FixMatch, except we reduce the total optimization iterations by half.

For experiments on ImageNet-1K, we use ResNet50 as the backbone network and follow the training recipes introduced in FixMatch w/ EMAN \cite{cai2021exponential}, which is also the default baseline of all experiments on ImageNet-1K. The model is initialized with MoCo v2 + EMAN as in \cite{cai2021exponential}. For the setting with multiple views, we perform two strong augmentations and two weak augmentations on each unlabeled sample. Each strongly-augmented instance is paired with one weakly-augmented instance, and we jointly optimize the two pairs via pseudo-labeling as in the original setting of Fig.~\ref{fig:Diagram}. Multi-views could increase the convergence speed and stabilize the training process.

\noindent\textbf{DebiasPL is simple yet effective}. 
Tab.~\ref{table:cifar} and Tab.~\ref{table:imagenet} show that DebiasPL delivers state-of-the-art performance on all experimented benchmarks, outperforming current approaches by a large margin. 
Without using CLIP, DebiasPL can outperform CoMatch on CIFAR, and is comparable to CoMatch on ImageNet-1K. DebiasPL wins on its merit of simplicity. Leveraging the power of CLIP could significantly improve the performance of DebiasPL, surpassing CoMatch by about 4\% on ImageNet-1K SSL.

\tabCIFARDist{ht}
\tabAblationSSLMethods{ht}

\noindent\textbf{DebiasPL is agnostic to source/target data distribution}. Tab.~\ref{table:cifar} shows that, for both CIFAR and long-tailed CIFAR SSL benchmarks, using a unified framework and the same set of hyper-parameters, DebiasPL can surpass previous state-of-the-art methods, which are either designed for balanced data or meticulously tuned for long-tailed data. Furthermore, Tab.~\ref{table:cifar-dist} shows that when tested in scenarios where labeled and unlabeled data follow different distributions, DebiasPL produces an even greater gain (11.4\%) to the baseline. 

\noindent\textbf{The fewer labeled data, the more significant gains} can be observed in Tab.~\ref{table:cifar} and Tab.~\ref{table:imagenet}, almost eliminating the gap between fully-supervised and semi-supervised learning.

\noindent\textbf{DebiasPL is also a universal add-on} as illustrated in Tab.~\ref{table:methods-SSL}.
Incorporating DebiasPL into various SSL methods can achieve consistent performance improvements.

\tabImageNetZeroShot{!t}

\figAblationZeroShotDatasets{!t}

\subsection{Tranductive Zero-Shot Learning}
\noindent\textbf{Dataset}. We evaluate the efficiency of DebiasPL in T-ZSL on ImageNet-1K \cite{ILSVRC15}. 
EuroSAT \cite{helber2019eurosat}, MNIST \cite{lecun1998mnist}, CIFAR10 \cite{krizhevsky2009learning}, CIFAR100 \cite{krizhevsky2009learning}, and Food101 \cite{bossard14} are also used as evaluation datasets to show the robustness to domain shift.

\noindent\textbf{Setup}. 
T-ZSL assumes that the list of possible class candidates is known for the target data.
Following this setting, we do not use any semantic labels for target data. 
We apply DebiasPL on CLIP in a similar way as we apply DebiasPL on FixMatch, except that the labeled data is ``labeled" by CLIP rather than a human annotator. Specifically, all unlabeled instances whose CLIP confidence score greater than $\tau_{\text{clip}}$ are pseudo-labeled by CLIP and considered as ``labeled" data.
A backbone of ResNet50 and a threshold $\tau_{\text{clip}}$ of 0.95 are used for all datasets. 
The same default hyper-parameters and training recipes as in FixMatch + EMAN are utilized for fair comparisons. 
More details are in the appendix.

\noindent\textbf{DebiasPL delivers SOTA results on zero-shot learning}, even surpassing CLIP \cite{radford2021learning} and CoOP \cite{zhou2021learning} that are fine-tuned on partial human-labeled data. Moreover, DebiasPL with a backbone of ResNet50 can significantly outperform CLIP with 15$\times$ larger backbones, as shown in Tab.~\ref{table:zeroshot}. The time cost of zero-shot training DebiasPL w/ CLIP (without using any human annotations) for 100 epochs is less than 0.01\% of CLIP's overall training time.

\noindent\textbf{DebiasPL exhibits stronger robustness to domain shift} than zero-shot CLIP without accessing any semantic labels, as depicted in ~\fig{datasets-ZSL}. Also, DebiasPL can observe greater gains (more than 20\%) on datasets with larger domain shifts, e.g., an astonishing 25.7\% gains can be obtained on the satellite image dataset EuroSAT \cite{helber2019eurosat}. 
\section{Summary}
In this paper, we conduct research on the previously unknown biased pseudo-labeling issue. A simple yet effective method DebiasPL is proposed to dynamically alleviate biased pseudo-labels’ influence on a student model, without leveraging any prior knowledge of true data distribution. As a universal add-on, DebiasPL delivers significantly better performance than previous state-of-the-arts on both semi-supervised learning and transductive zero-shot learning tasks and exhibits stronger robustness to domain shifts. 

\noindent{\bf Acknowledgements.} 
This work was supported, in part, by US Government fund through Etegent Technologies on Low-Shot Detection and Semi-supervised Detection.

\begin{algorithm*}[t]
\footnotesize
\SetAlgoLined
    \PyComment{\scriptsize initialize p\_hat with 1/C, C is the number of classes} \\
    \PyCode{p\_hat = torch.ones([1, C]) / C} \\
    \PyComment{\scriptsize load a batch with unlabeled and labeled samples} \\
    \PyComment{\scriptsize x: labeled samples ; target: labels for x ; u: unlabeled samples} \\
    \PyCode{for (x, target), u in loader:} \\
    \Indp   
        \PyComment{\scriptsize augment x with weak augmentation and get two versions of u with strong and weak augmentations} \\
        \PyCode{x, u\_s, u\_w = {weak}(x), {strong}(u), {weak}(u)} \\
        \PyComment{\scriptsize model forward} \\
        \PyCode{l\_x, l\_us, l\_uw = model(x, u\_s, u\_w)} \\ 
        \vspace{6pt}
        
        \PyComment{\scriptsize get debiased pseudo-labels} \\
        p\_uw = \PyCode{F.softmax(l\_uw - tau * torch.log(p\_hat), dim=1)} \\ 
        \PyCode{max\_probs, pseudo\_label = torch.max(p\_uw, dim=-1)} \\
        \vspace{6pt}

        \PyComment{\scriptsize get mask for filtering instances with low confidence score} \\
        \PyCode{mask = max\_probs.ge(thresh).float()} \\
        \PyComment{\scriptsize update p\_hat} \\
        p\_hat = momentum * p\_hat + (1 - momentum) * p\_uw.detach().mean(dim=0) \\
        \vspace{6pt}
        
        \PyComment{\scriptsize calculate loss\_x for labeled instances} \\
        \PyCode{loss\_x = F.cross\_entropy(l\_x, target)} \\
        \PyComment{\scriptsize calculate marginal loss loss\_u for unlabeled instances} \\
        \PyCode{l\_us = l\_us + lambda * torch.log(p\_hat)} \\
        \PyCode{loss\_u = (F.cross\_entropy(l\_us, pseudo\_label, reduction='none') * mask).mean()} \\
        \PyComment{\scriptsize total loss} \\
        \PyCode{loss = loss\_x + lambda\_u * loss\_u} \\
        \vspace{6pt}
        
        \PyComment{\scriptsize optimization step} \\
        \PyCode{loss.backward()} \\
        \PyCode{optimizer.step()} \\
        
    \Indm 
    \PyComment{\scriptsize update the ema model} \\
    \PyCode{model.momentum\_update\_ema()}
\caption{\small PyTorch-style pseudocode for semi-supervised learning with DebiasPL}
\label{algo:pseudo-code}
\end{algorithm*}

\def\tabAblationCifarLambda#1{
\begin{table}[#1]
\tablestyle{6pt}{1.0}
\small
\centering
\begin{tabular}{l||cccccc}
\shline
{$\lambda$} & 0.0 & 0.25 & 0.5 & 0.75 & 1.0 & 2.0 \\
\shline
DebiasPL & 73.5 & 79.5 & 80.6 & 80.5 & 80.5 & 77.7 \\
\shline
\end{tabular}\vspace{-4pt}
\caption{\textbf{Ablation study} on CIFAR10-LT ($\gamma\!=\!100$) semi-supervised learning with DebiasPL under various \textbf{weight $\lambda$} of debiasing module and marginal loss. 30\% samples are labeled. The model is identical to FixMatch when $\lambda=0$. Results averaged over 5 different folds are reported.
}
\label{table:tau-ablation}
\end{table}
}

\def\tabAblationCifarComponent#1{
\begin{table}[#1]
\tablestyle{6.5pt}{1.0}
\small
\centering
\begin{tabular}{cc||cc}
\shline
Debiasing & Magirnal Loss & CIFAR10 & CIFAR10-LT \\ 
\shline
 &  & 86.1 & 73.5 \\
 \hline
$\cmark$ &  & 93.3 & 79.6 \\
$\cmark$ & $\cmark$ & 94.6 & 80.6 \\
\shline
\end{tabular}\vspace{-4pt}
\caption{\textbf{Ablation study} on the \textbf{contribution of each component} of DebiasPL. Experimented on CIFAR10 and CIFAR10-LT ($\gamma\!=\!100$) SSL, in which 4 out of 5,000 samples are labeled per class for CIFAR10 and 30\% instances are labeled for CIFAR10-LT. Results averaged over 5 different folds are reported.
}
\label{table:component-ablation}
\end{table}
}

\def\tabAblationImageNetZeroShot#1{
\begin{table}[#1]
\tablestyle{4pt}{1.0}
\small
\centering
\begin{tabular}{l||cccccc}
\shline
{$\tau_{\text{clip}}$} & 0.2 & 0.4 & 0.6 & 0.8 & 0.9 & 0.95 \\ 
\shline
DebiasPL + CLIP & 55.9 & 63.2 & 66.2 & 67.1 & 67.7 & 67.7 \\
\shline
\end{tabular}\vspace{-4pt}
\caption{\textbf{Ablation study} on ImageNet-1K zero-shot Learning with DebiasPL + CLIP \cite{radford2021learning} under various \textbf{threshold $\tau_{\text{clip}}$}. 
}
\label{table:zeroshot-ablation}
\end{table}
}

\def\figCLIPZSL#1{
    \captionsetup[sub]{font=small}
    \begin{figure*}[#1]
      \centering
      \includegraphics[width=0.9\linewidth]{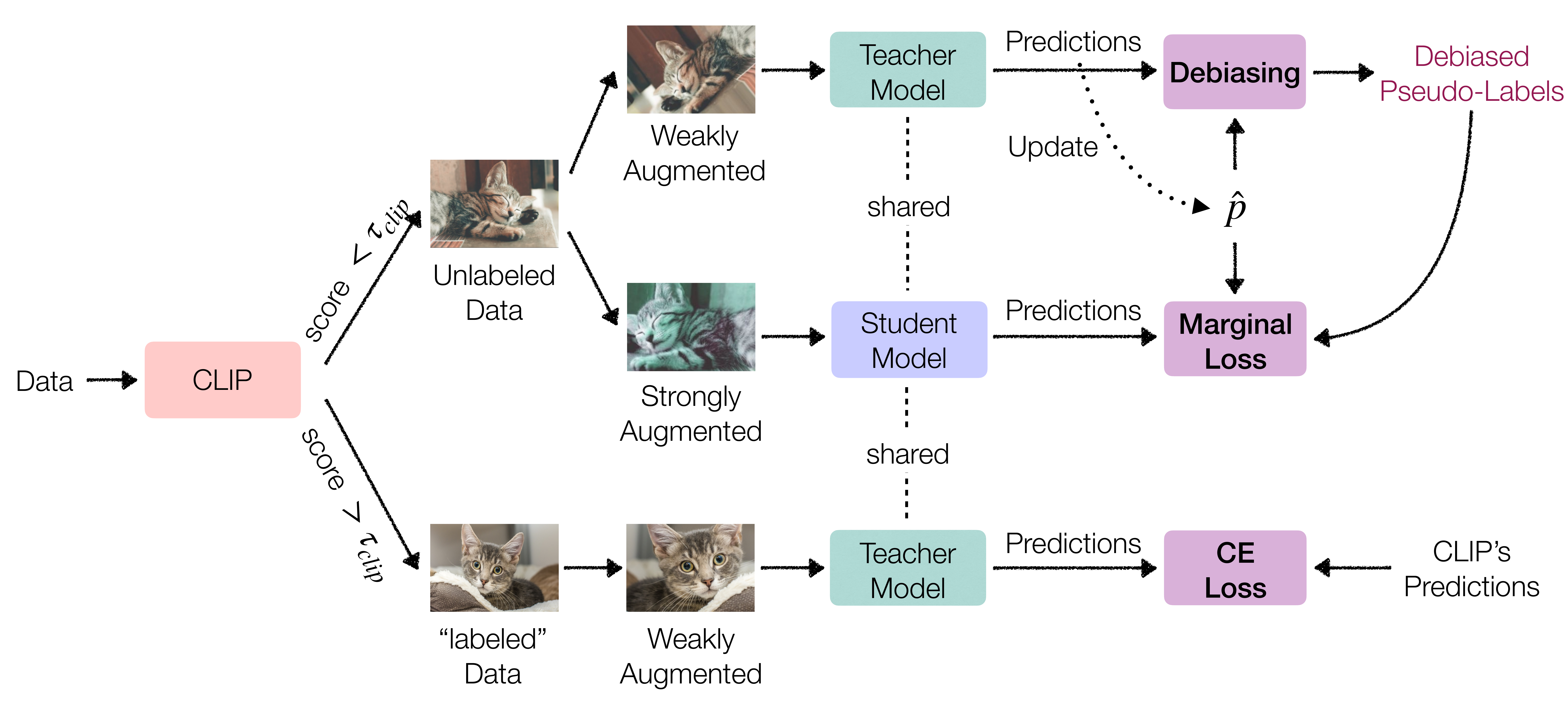}\vspace{-4pt}
      \caption{The overall framework of transductive zero-shot learning with CLIP + DebiasPL. CLIP + FixMatch can be realized by removing the debiasing module and replacing the marginal loss with cross-entropy loss.}
      \label{fig:framework-zsl}
    \end{figure*}
}

\def\figCLIP#1{
    \captionsetup[sub]{font=small}
    \begin{figure}[#1]
      \centering
      \begin{subfigure}{0.32\linewidth}
        \includegraphics[width=1.0\linewidth]{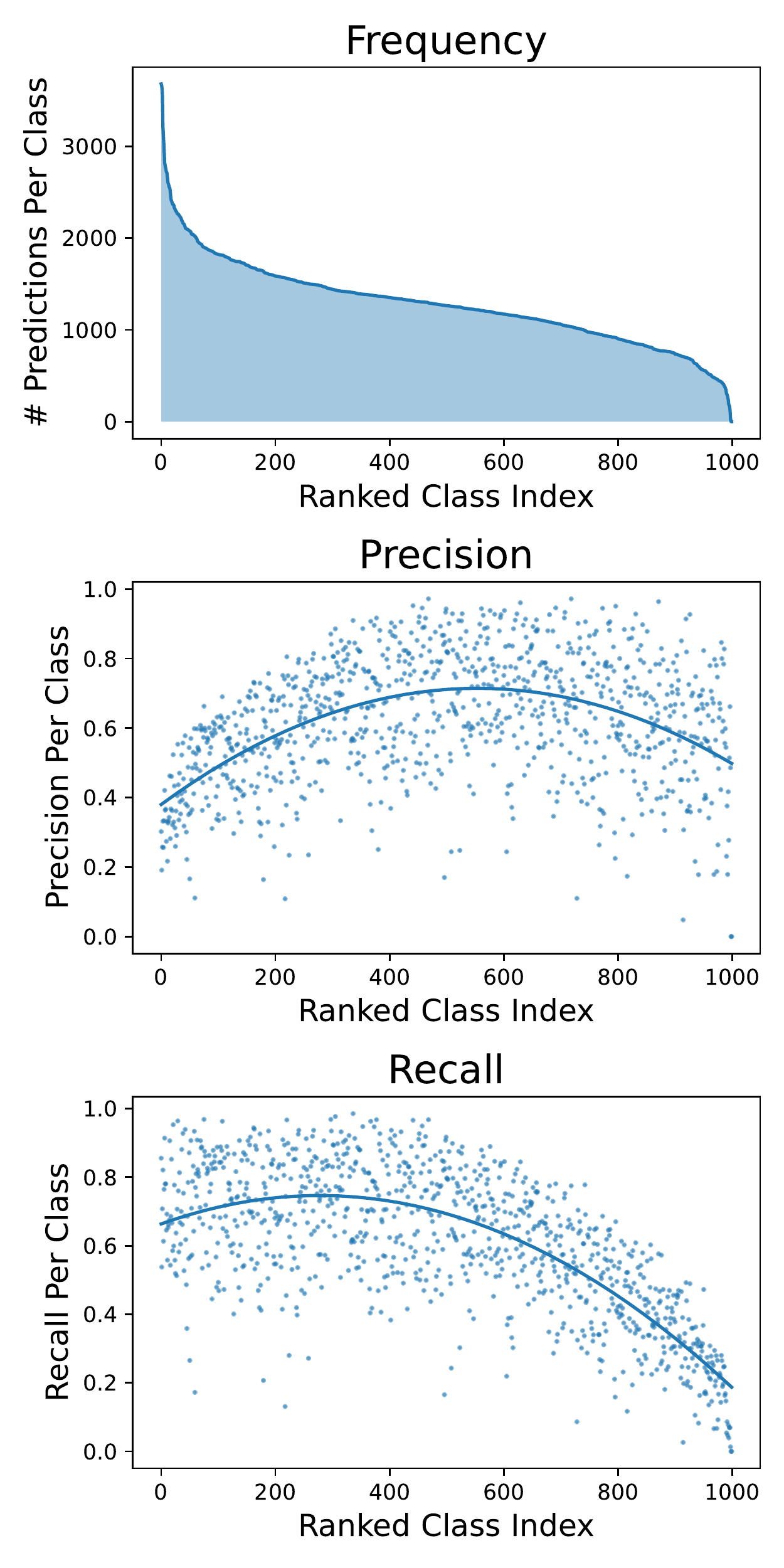}
        \caption{$\tau$ $\geq$ 0.0}
        \label{fig:short-a}
      \end{subfigure}
      \hfill
      \begin{subfigure}{0.32\linewidth}
        \includegraphics[width=1.0\linewidth]{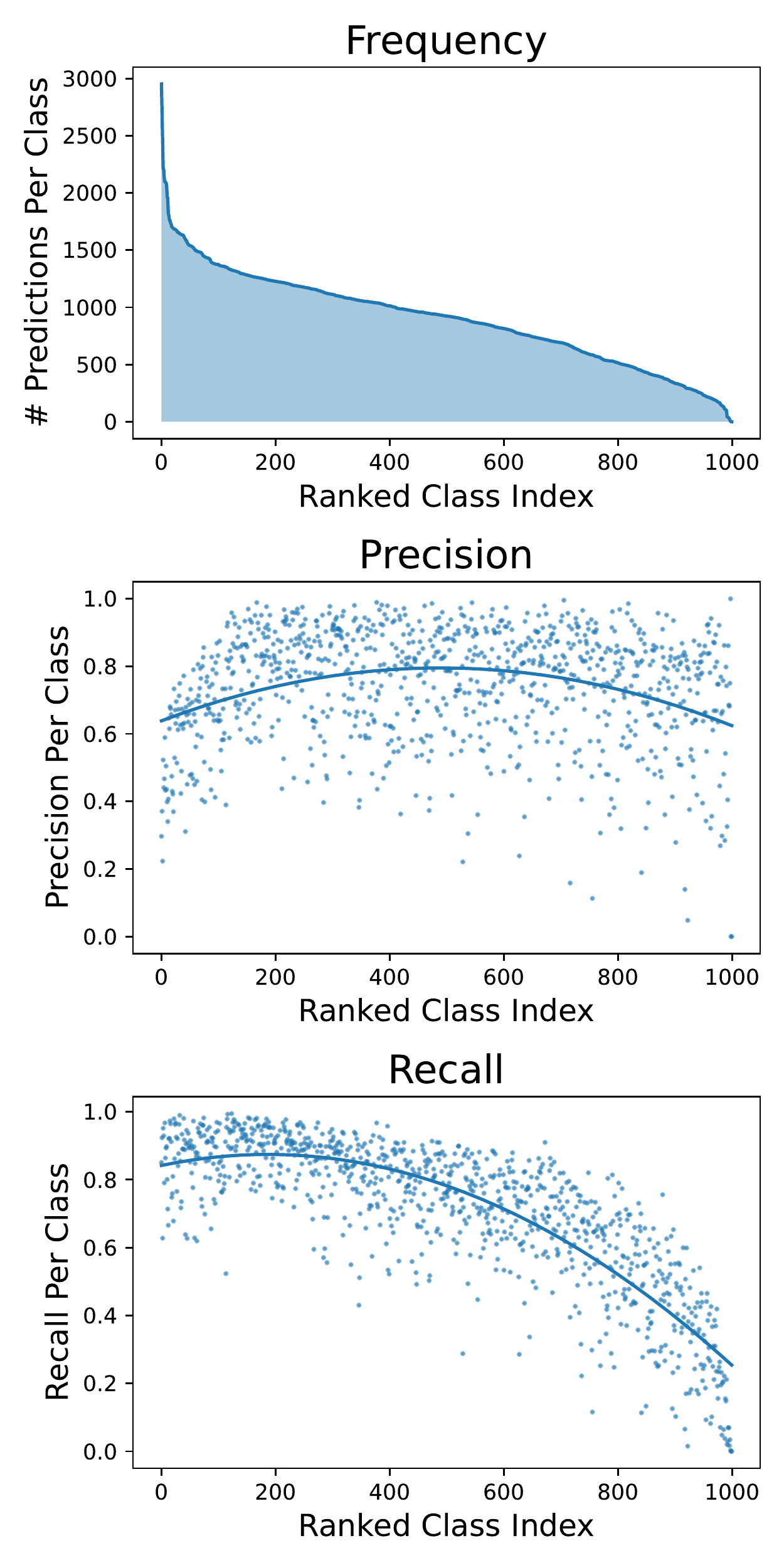}
        \caption{$\tau$ $\geq$ 0.4}
        \label{fig:short-a}
      \end{subfigure}
      \hfill
      \begin{subfigure}{0.32\linewidth}
        \includegraphics[width=1.0\linewidth]{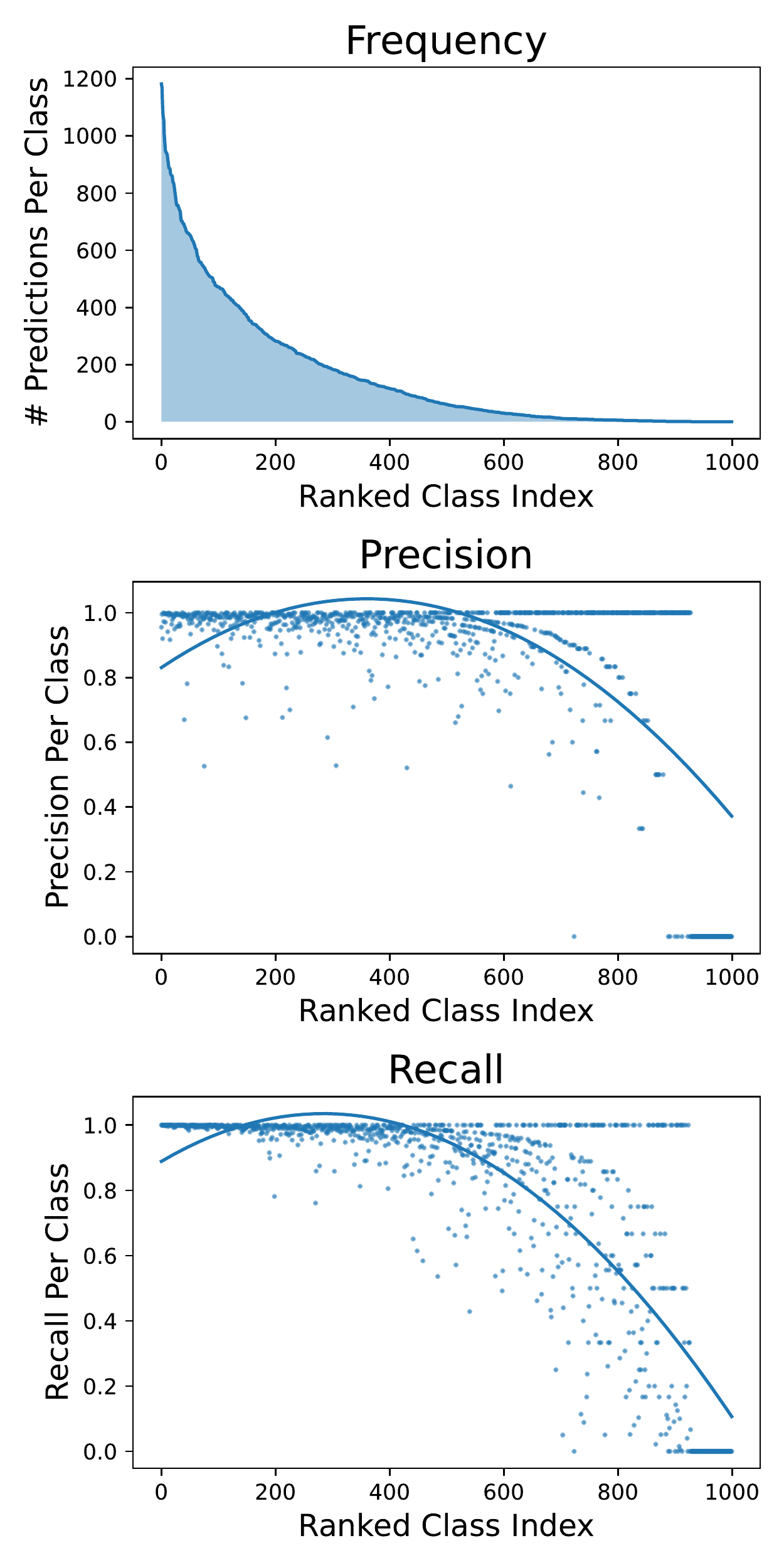}
        \caption{$\tau$ $\geq$ 0.95}
        \label{fig:short-b}
      \end{subfigure}
      \caption{A higher imbalanced ratio is obtained when filtering CLIP's zero-shot predictions with a larger threshold, analyzed on CLIP's zero-shot predictions on 1.3M almost class-balanced ImageNet training samples. 
      Per class number of predictions (row 1), precision (row 2), and recall (row 3) of samples passing various confidence score thresholds $\tau$ are visualized. Zero-shot predictions are produced with an ensemble of 80 prompts and a backbone of ResNet50, using official codes.}
      \label{fig:clip}
    \end{figure}
}


\section{Appendix}
\subsection{Details on Datasets and Implementations}
The PyTorch-style pseudocode for semi-supervised learning with DebiasPL is available at  Algo.~\ref{algo:pseudo-code}.

We conduct experiments on several benchmarks to prove the effectiveness and universality of DebiasPL. Here we provide more details on datasets and implementations for each benchmark:

{\bf{CIFAR10}} \cite{krizhevsky2009learning}: The original version of CIFAR10 contains 50,000 images on the training set and 10,000 images on the validation set with 10 categories for CIFAR10. 
For semi-supervised learning on CIFAR10, we conduct the experiments with a varying number of labeled examples from 40 to 250, following standard practice in previous works \cite{berthelot2019mixmatch, berthelot2019remixmatch, sohn2020fixmatch, li2021comatch}.
The reported results of each previous method in the paper are directly copied from the best-reported results in MixMatch \cite{berthelot2019mixmatch}, ReMixMatch \cite{berthelot2019remixmatch}, FixMatch \cite{sohn2020fixmatch}, CoMatch \cite{li2021comatch}, etc.

We keep all hyper-parameters the same as FixMatch, except for the number of training steps.
We use WideResNet-28-2 \cite{he2016deep, zagoruyko2016wide} with 1.5M parameters as a backbone network for CIFAR10. The SGD optimizer with a Nesterov momentum of 0.9 is used for optimization. The learning rate is initialized as 0.03 and decayed with a cosine learning rate scheduler \cite{loshchilov2016sgdr}, which sets the learning rate at training step $k$ as $\cos(\frac{7\pi k}{16K})$ times the initial learning rate, where $K=2^{19}$ is the total number of training steps, i.e., about 512 epochs, and is 2 times fewer than the original number of FixMatch training steps.
The model is trained with a mini-batch size of 512, which contains 64 labeled samples and 448 unlabeled samples, on one V100 GPU. As in previous works, an exponential moving average of model parameters is used to produce the final performance. The weight decay is set as 0.0005 for CIFAR10. 
Unless otherwise stated, the only independent hyperparameter of DebiasPL $\lambda$ is fixed and set to $0.5$ in all experiments.
Each method is tested under 5 different folds, and we report the mean and the standard deviation of accuracy on the test set.

{\bf{CIFAR10-LT}} \cite{krizhevsky2009learning, liu2019large, wei2021crest}: The long-tailed version of CIFAR10 follows an exponential decay in sample sizes across different categories. CIFAR10-LT is constructed by sampling a subset of CIFAR10 following the Pareto distribution with the power value $\gamma\in [100, 200]$.
Then, we select 10\% or 30\% of all CIFAR10-LT instances to construct the SSL benchmark labeled dataset, and the others are regarded as the unlabeled datasets.
Each algorithm is tested under 5 different folds of labeled data, and we report the mean and the standard deviation of accuracy on the test set. 
As in previous works, an exponential moving average of model parameters is used to produce the final performance.

To demonstrate the universality of the proposed method DebiasPL and its insensitivity to data distribution, we follow the same hyperparameters and training formulas in CIFAR10. We do not specifically adjust any hyperparameters when conducting experiments in the long-tail SSL benchmarks.

{\bf{ImageNet-1K}} \cite{ILSVRC15}: ImageNet-1K is a curated dataset with approximately class-balanced data distribution, containing about 1.3M images for training and 50K images for validation.  

For semi-supervised learning, ImageNet-1K with varying amounts of labeled data is experimented with, i.e., 0.2\% and 1\%. 
The FixMatch model is trained with a batch size of 64 (320) for labeled (unlabeled) images with an initial learning rate of 0.03. Following \cite{cai2021exponential}, we replace batch normalization (BN) layers with exponential moving average normalization (EMAN) layers in the teacher model. 
EMAN updates its statistics by exponential moving average from the BN statistics of the student model.
ResNet-50 is used as the default network and the default hyperparameters in the corresponding papers \cite{cai2021exponential, sohn2020fixmatch} are applied. 
The model is initialized with MoCo v2 + EMAN pre-trained model as in \cite{cai2021exponential}. 
To make fair comparisons, we report results of FixMatch with EMAN as the baseline model, and all hyper-parameters of FixMatch with EMAN are untouched unless noted otherwise. 

\figCLIPZSL{!t}
For zero-shot learning, no manual annotation is leveraged in the training process. 
We train CLIP + DebiasPL and CLIP + FixMatch following the same hyperparameters and training recipes as FixMatch with EMAN, except that the labeled data is ``labeled" by CLIP rather than a human annotator. Specifically, all unlabeled instances whose CLIP confidence score greater than $\tau_{\text{clip}}$ are pseudo-labeled by CLIP (with a backbone of ResNet50) and considered as ``labeled" data.
A backbone of ResNet50 and a threshold $\tau_{\text{clip}}$ of 0.95 are used. 
The same default hyper-parameters and training recipes as in FixMatch + EMAN are utilized for fair comparisons. The framework of transductive zero-shot learning with DebiasPL is illustrated in Fig.~\ref{fig:framework-zsl}.

For experiments on other benchmarks of ZSL, including EuroSAT \cite{helber2019eurosat}, MNIST \cite{lecun1998mnist}, DTD \cite{cimpoi2014describing}, GTSRB \cite{stallkamp2011german} and Flowers102 \cite{nilsback2008automated}, we follow the training recipe of ImageNet-1K.

\subsection{Ablation Study}
In this section, we conduct additional ablation studies on the influence of the two components of DebiasPL (Table.~\ref{table:component-ablation}) for SSL, DebiasPL’s unique hyperparameter $\lambda$ (Table.~\ref{table:tau-ablation}) for SSL, and CLIP’s confidence score threshold $\tau_{\text{clip}}$ (Table.~\ref{table:zeroshot-ablation}) for T-ZSL. 

\tabAblationCifarComponent{ht}
As shown in Table.~\ref{table:component-ablation}, the two components of DebiasPL lead to significant improvements to \textit{both} CIFAR10 and CIFAR10-LT SSL benchmarks. 
Compared with the balanced benchmark, the performance improvement obtained by introducing the marginal loss is relatively smaller than the unbalanced benchmark.

\tabAblationCifarLambda{ht}
Table.~\ref{table:tau-ablation} illustrates the influence of debias factor $\lambda$. When the value of $\lambda$ is set to 0, DebiasPL is identical to FixMatch. Adding a debiasing module and marginal loss can improve the performance on CIFAR10-LT by more than 7\% when selecting the optimal choice of $\lambda$ 0.5, which is marginally better than the default value of 1.0. However, there is a trade-off. Suppose the debias factor $\lambda$ is too strong. In that case, it is hard for a model to fit on the data, while a too-small factor can barely eliminate the biases, ultimately impairs the generalization ability. 

\figCLIP{ht}
\tabAblationImageNetZeroShot{ht}
As illustrated in the main paper, the CLIP predictions are class-imbalanced. Therefore, the natural question is whether we can obtain a more balanced prediction by filtering instances with a threshold $\tau_{\text{clip}}$? 
Unfortunately, no, on the contrary, when filtering predictions with a larger threshold, a higher imbalance rate is observed, as in \fig{clip}.
Furthermore, when filtering instances with a threshold of 0.95, more than 60 categories get zero predictions. 

The dilemma is that using a smaller threshold $\tau_{\text{clip}}$ can obtain a smaller imbalanced ratio, which is the desired property. However, it also leads to a lower precision, introducing many outliers and misclassified samples. 
Therefore, a module to eliminate biases captured by the CLIP model when CLIP is pre-trained on source data is needed to yield a good performance on target data. 

Table.~\ref{table:zeroshot-ablation} shows that using a threshold of 0.95 can get the optimal performance on the ImageNet zero-shot learning task, which indicates that the high precision of the labeled data, realized by using a high threshold, is essential for better performance on target data. At the same time, our proposed DebiasPL can greatly alleviate the trouble of a higher imbalance ratio caused by using a larger threshold, eventually obtaining more than 10\% performance gains.

{\small
\bibliographystyle{ieee_fullname}
\bibliography{egbib}
}

\end{document}